\documentclass{article}
\usepackage{graphicx} 
\usepackage{xcolor}
\usepackage{comment}
\usepackage{booktabs}
\usepackage{pdfpages}
\usepackage[acronym]{glossaries}
\usepackage{siunitx}
\usepackage{geometry}
\usepackage{float}
\usepackage{hyperref}
\usepackage{xr-hyper}
\usepackage{xr}
\newacronym{spa}{SPA}{soft pneumatic actuator}
\newacronym{sota}{SOTA}{state-of-the-art}
\newacronym{fem}{FEM}{finite element method}
\newacronym{prbm}{PRBM}{pseudo-rigid body model}
\newacronym{pcc}{PCC}{piecewise constant curvature}
\newacronym{mor}{MOR}{model order reduction}
\newacronym{pod}{POD}{proper orthogonal decomposition}
\newacronym{mse}{MSE}{mean squared error}
\newacronym{rmse}{RMSE}{root-mean-squared error}
\newacronym{ppo}{PPO}{Proximal Policy Optimization}
\newacronym{nn}{NN}{neural network}
\newacronym{cma-es}{CMA-ES}{Covariance Matrix Adaptation Evolution Strategy}
\newacronym{rl}{RL}{Reinforcement Learning}
\newacronym{xpbd}{XPBD}{extended position-based dynamics}
\newacronym{mpm}{MPM}{material point methods}

\title{Generalized Task-Driven Design of Soft Robots via Reduced-Order FEM-based Surrogate Modeling}

\author{
Yao Yao\\
Oxford Robotics Institute\\ University of Oxford, UK
\and
David Howard\\
CSIRO Robotics\\ CSIRO, Australia
\and
Perla Maiolino\thanks{Corresponding author: perla.maiolino@eng.ox.ac.uk}\\
Oxford Robotics Institute\\ University of Oxford, UK
}

\date{}

\begin{document}
\maketitle

\begin{abstract}

Task-driven design of soft robots requires models that are physically accurate and computationally efficient, while remaining transferable across actuator designs and task scenarios. However, existing modeling approaches typically face a fundamental trade-off between physical fidelity and computational efficiency, which limits model reuse across design and task variations and constrains scalable task-driven optimization.
This paper presents a unified reduced-order \gls{fem}-based surrogate modeling pipeline for generalized task-driven soft robot design. High-fidelity \gls{fem} simulations characterize actuator behavior at the modular level, from which compact surrogate joint models are constructed for evaluation within a \gls{prbm}. A meta-model maps actuator design parameters to surrogate representations, enabling rapid instantiation across a parameterized actuator family. The resulting models are embedded into a \gls{prbm}-based simulation environment, supporting task-level simulation and optimization under realistic physical constraints.
The proposed pipeline is validated through sim-to-real transfer across multiple actuator types, including bellow-type pneumatic actuators and a tendon-driven soft finger, as well as two task-driven design studies: soft gripper co-design via \gls{rl} and 3D actuator shape matching via evolutionary optimization. The results demonstrate high accuracy, efficiency, and reliable reuse, providing a scalable foundation for autonomous task-driven soft robot design.

\end{abstract}

\section{Introduction}
Soft robots are increasingly valued for their adaptability, compliance, and ability to achieve a wide range of behaviors through the combination of material properties, morphology, and control~\cite{Rus2015, Lee2017, Shintake2018}. 
However, these advantages also make the design of soft robots a significant challenge, leading to vast and complex design spaces where manual or heuristic approaches are typically insufficient to ensure optimal performance in many task applications. This has led to an increasing focus on task-driven design, wherein morphology and control are systematically tailored to meet specific application objectives via computational approaches~\cite{Pinskier2022, Stella2023}.

Such approaches rely heavily on accurate and scalable modeling to explore large design spaces. However, soft systems pose particular difficulties in this context, especially when predicting forces and self or environmental interactions, due to their inherent softness and material nonlinearity. 
This not only challenges the commonly discussed issue of accuracy and efficiency, but also reveals a critical modeling bottleneck: the lack of generalizable models that can be reused across parameterized geometries, actuation types, and task scenarios. Here, generalization refers to the ability of a model to retain accuracy across variations within a parameterized design space and across different task-level conditions represented in simulation, rather than being confined to case-specific configurations.


This limitation stems from a fundamental trade-off in modeling soft robots.  \textcolor{black}{Despite a wide variety of modeling techniques having been explored, including first-principles-based methods, physics-based simulation, and data-driven or reduced-order models, most approaches follow one of two dominant directions: high-fidelity modeling and simplified or reduced-order representations.}
Approaches based on high-fidelity modeling, such as \gls{fem}, offer accurate predictions but are often too computationally expensive for repeated task-level evaluation, which in practice encourages the use of simplified or proxy objectives tied to specific tasks and assumptions~\cite{Pozzi2018, Xavier2021}.
In contrast, simplified or reduced-order models are computationally efficient, but typically rely on assumptions and approximations for particular designs or operating conditions, limiting their transferability to new geometries, loading regimes, or task settings~\cite{Schaff2023, Navez2025}. 

As a result, existing approaches fall short of providing models that are simultaneously accurate and efficient while remaining sufficiently general for task-driven design.
This lack of generalization limits model scalability and deployability across broader design contexts, thereby hindering the development of general-purpose modeling tools that the research community can readily build upon. Addressing this limitation is essential for building flexible and reusable design pipelines in soft robotics.

To overcome this challenge, this paper introduces a unified modeling pipeline for task-driven design of soft robots. Specifically, we:
\begin{itemize}
    \item Use high-fidelity \gls{fem} simulations to generate actuator data that captures coupled relationships  between actuation, deformation, and force at the modular level;
    \item Create surrogate models from this data using either polynomial regressions or small \gls{nn}s, serving as joint models within the \gls{prbm} framework, to \textcolor{black}{significantly accelerate simulation while maintaining \gls{fem}-level accuracy;}
    \item Develop a meta-model that maps actuator design parameters to surrogate features, enabling rapid model interpolation within a parameterized actuator family;
    \item Integrate the resulting models into a \textcolor{black}{fast and flexible} \gls{prbm}-based simulation environment, such as PyBullet, supporting efficient task-level simulation and optimization under real-world physical constraints such as contact, loading, and gravity.
\end{itemize}
Together, these contributions support pipeline \textcolor{black}{reuse across design variations and task conditions,} offering a practical pathway toward scalable task-driven design. Notably, the proposed approach is scoped to parameterized actuator families and is not intended to generalize across fundamentally different soft robotic morphologies or actuation principles.

We illustrate our contributions using a range of different soft actuators, including pneumatically actuated 
\textcolor{black}{bellow-type soft actuators, which are widely used in soft robotic field due to their large deformation, adaptability, straightforward control, and 3D printing compatibility. These actuators remain challenging to model accurately while efficiently due to effectively infinite degrees of freedom and distributed actuation~\cite{SPAreview}.}
We also include a tendon-driven finger, demonstrating that the proposed pipeline extends beyond a single actuator type. Across diverse applications including soft gripping and 3D shape-matching, we demonstrate that the pipeline supports both accurate physical modeling and efficient task-level optimization. An overview of the workflow is provided in Figure~\ref{fig:pipeline}.
\section{Literature Review}
\label{sec:LR}
Modeling plays a critical role in the design of soft robots, as it directly impacts both the accuracy of the predicted behavior and the speed of the automated design process. A wide range of modeling approaches is available, each with its own trade-offs between accuracy, efficiency, and generalizability.

Analytical models, such as rod formulations~\cite{alessi2024rodmodelscontinuumsoft} or minimum potential
energy principle~\cite{Bruder2023}, are valued for their simplicity and computational efficiency. However, they often rely on simplifying assumptions, such as uniform curvature, constant cross-sections, or linearized material behavior. These assumptions limit flexibility and often require re-derivation or re-calibration when designs, materials, or boundary conditions change.

Geometric models, such as Piecewise Constant Curvature (PCC) and function-based curves, provide efficient representations of the kinematics of soft actuators and intuitive mappings from actuation inputs to robot configurations~\cite{WebsterIII2010}. However, they tend to break down under unexpected external forces such as gravity or contact~\cite{Navez2025}.

Discrete models, such as \gls{prbm}s, approximate continuum deformation by connecting rigid segments with compliant joints~\cite{PRBM-reviewer2}. They are lightweight and compatible with existing rigid-body simulators (e.g., PyBullet), enabling reliable and efficient task- and environment-level simulation~\cite{Cosimo2018, SoMo2021}. However, they typically require careful parameter calibration to map the lumped behavior back to distributed soft structures. This process is time-consuming and often lacks generalizability beyond the specific robot for which the model is tuned. 
Such limitations highlight the need for modeling approaches that can be reused across actuator designs, enabling broader applicability within the research community.

Numerical methods, especially \gls{fem}, offer the highest fidelity by solving governing equations over the continuum. 
\gls{fem} is widely used for simulating soft robots because it provides high accuracy and strong generality, accommodating arbitrary geometries, materials, actuation, and boundary conditions, with established tool support such as COMSOL Multiphysics\textsuperscript{\textregistered}, ANSYS\textsuperscript{\textregistered}, and Abaqus\textsuperscript{\textregistered}~\cite{Moseley2016, Xavier2021, Ferrentino2024}. 
\textcolor{black}{Other numerical approaches have also been explored for deformable simulation, particularly in the computer graphics community, including \gls{xpbd}~\cite{XPBD} and \gls{mpm}~\cite{MPM}. \gls{xpbd} enables efficient and stable simulations but often yields less reliable mechanical responses, while \gls{mpm} can capture large deformations but may introduce numerical diffusion that affects accurate force responses.}
However, the high computational cost and convergence challenges of \gls{fem} under complex contact and large-deformation conditions limit its applicability in large-scale design optimization.
As a result, task-level performance is often approximated through actuator-level objectives that are more tractable to evaluate but are closely tied to specific modeling assumptions and task formulations.
This limitation has motivated efforts in model order reduction techniques, which seek to retain fidelity while reducing computational expense~\cite{Pozzi2018, Wiese2022}.

\Gls{mor} techniques aim to accelerate \gls{fem} by reducing system dimensionality while retaining dominant deformation modes. Examples include \gls{pod}~\cite{Goury2018, Schaff2023}, condensed \gls{fem} formulations~\cite{Navez2025}, and Cosserat rod implementations such as SoRoSim~\cite{SoRoSim2023}. These methods improve computational efficiency and can reproduce complex deformations with fewer degrees of freedom.
However, the reduced bases or governing formulations are typically derived for specific geometries, boundary conditions, or actuation schemes, making the resulting models configuration-specific and difficult to reuse in task-driven design workflows.

Surrogate models learn input–output relationships from high-fidelity simulations or experiments, providing fast evaluation once trained. Approaches include regression models, Gaussian processes, and \gls{nn}s that map actuation to deformation and force responses~\cite{Liu2022, Yao2024}. Compared with physics-based reductions, data-driven surrogates can offer greater flexibility. However, their performance is inherently tied to the coverage of the training data, trained under fixed loading conditions or tailored to a specific design, making it hard to extrapolate reliably when tasks, geometries, or operating regimes change.

In summary, existing approaches present clear trade-offs. Analytical, geometric, and discrete models are efficient but rely on restrictive assumptions that limit accuracy and transferability. \gls{fem} provides accuracy and generality across arbitrary geometries and boundary conditions, but is too costly to support large-scale task-level design optimization. Physics-based model reduction improves efficiency but remains tied to specific designs or loading conditions. Data-driven surrogates accelerate evaluation and can capture complex behaviors, yet they are often restricted to fixed training scenarios and lack generalization across tasks. Overall, what is still missing is a pipeline that unifies these ideas into models that are accurate, efficient, and reusable across designs and task scenarios. Addressing this gap is essential not only for advancing task-driven design of soft robots but also for providing a modeling foundation that can be extended and adopted by the broader soft robotics community.

\section{Materials and Methods}
\label{sec:M&M}
\begin{figure}[t]
\includegraphics[width=\linewidth]{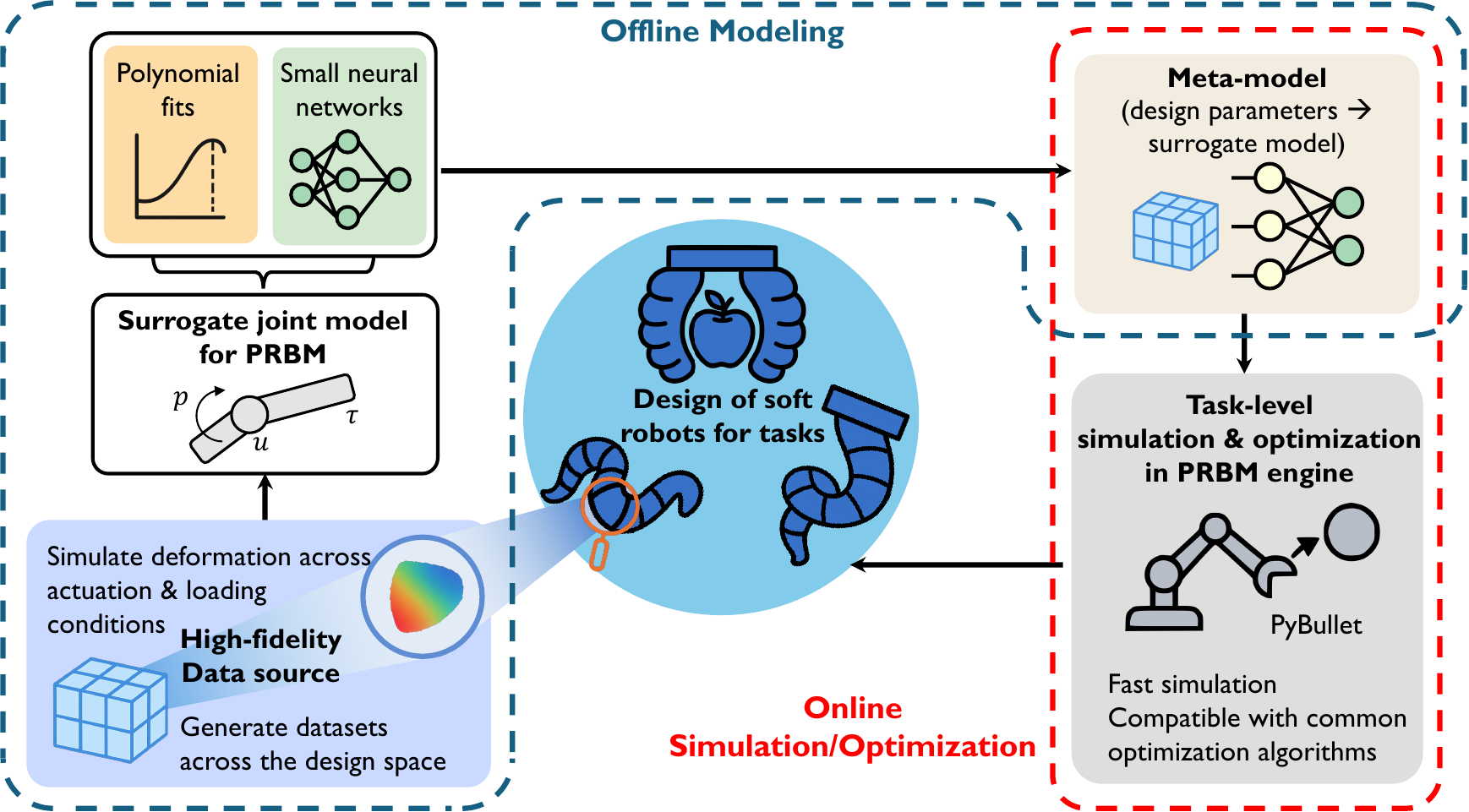}
  \caption{Schematic overview of the proposed pipeline for modeling and task-driven design of soft robots. The process starts from a well-parameterized modular actuator design, which is simulated using \gls{fem} to generate high-fidelity training data. Compact surrogate models are then constructed from these data as joint models, represented using either polynomial fits or small \gls{nn}s, for use in \gls{prbm}-based simulations. By repeating this process across different designs, meta-models are constructed to capture the mapping from design parameters to surrogate models. The resulting surrogate and meta-models can be integrated with \gls{prbm}-based simulators and \gls{sota} optimization algorithms, enabling fast task-level simulation and task-driven design optimization.}
\label{fig:pipeline}
\end{figure}

This section introduces the proposed modeling pipeline for task-driven design of soft robots. The pipeline combines four components: 
(i) high-fidelity \gls{fem} simulations to capture coupled actuation–deformation–force responses;
(ii) surrogate models that encode these behaviors into compact representations for fast evaluation, and (iii) \gls{prbm}-based simulation in PyBullet, embedding surrogates for task-level simulation. 
(iv) meta-models that map design parameters to surrogate models, supporting scalability across new actuator designs.

To explicitly illustrate this workflow, as presented in Figure~\ref{fig:pipeline}, the pipeline starts from a parameterized modular actuator design, whose behavior is first characterized using high-fidelity \gls{fem} under both free-loading and constrained conditions. 
Building on the \gls{fem} data, compact surrogate joint models are derived to approximate the nonlinear relationship between actuation input and mechanical response at the modular level for use within a \gls{prbm}-based simulation.
These surrogates significantly reduce computational cost while preserving physical fidelity.

To support design generalization, the surrogate construction process is repeated across multiple actuator designs, and the resulting model parameters are learned by a higher-level meta-model that maps from design parameters to surrogate representations. 
This enables rapid on-demand generation of behavior models for \textcolor{black}{unseen designs via interpolation within a parameterized actuator family}, eliminating the need for repeated \gls{fem} simulations. 
Finally, the design-conditioned surrogate models are embedded into \gls{prbm}-based simulation in PyBullet, enabling task-level simulation with realistic physical interactions. This integrated environment also supports efficient task-level design using \gls{sota} learning and optimization algorithms.

\subsection{\gls{fem} Data Generation}\label{sec:fem_data}
Existing surrogate models are often developed for specific loading conditions, such as free deformation, and may not capture the coupled effects of actuation and external forces. However, task-level simulations within a \gls{prbm} framework require joint laws that reflect both deformation and the corresponding forces. To address this, we design \gls{fem} simulations that provide a comprehensive mapping between actuation input, deformation, and reaction forces at the modular level.

\textcolor{black}{The \gls{fem} simulations are conducted under quasi-static assumptions to characterize the dominant elastic response of the actuator modules. This is appropriate for the considered systems, as the actuation and deformation processes occur at relatively low speeds, and inertial effects are negligible compared to elastic forces. The resulting actuation–deformation-force relationships are used as joint models, which are subsequently embedded into task-level simulations to account for gravity, contact, and external interactions.}

The procedure consists of two complementary conditions. In the free-loading condition, actuation inputs are incrementally applied to the actuator while external loads are absent, yielding the natural actuation–deformation response. In the constrained condition, the actuation input is held fixed while external forces are applied to vary the deformation, enabling explicit capture of the force–deformation relationship under constraint. 
\textcolor{black}{The \gls{fem} data are sampled over actuation ranges determined by the intended task operating conditions, while the external force range is constrained by stable FEM convergence. These simulations collectively yield coupled actuation–deformation–force data used to derive compact surrogate models.}

For clarity, an illustrative example using a translational bellow-\gls{spa} is provided in the Supplementary Material Section~1.1, where specific \gls{fem} configurations, parameter sweeps, and numerical results are detailed.

\subsection{Surrogate Model via Polynomial Fits}
\label{sec:surr_poly}
The \gls{fem} data described above provides actuator states under both free-loading and constrained conditions. To enable efficient task-level simulation within a \gls{prbm} framework, we derive a compact analytical surrogate that represents the joint force as a function of actuation and deformation. \textcolor{black}{Here $\tau$ denotes the generalized joint effort associated with the deformation coordinate $u$. Depending on the actuator configuration, this effort corresponds to a torque for rotational deformation or a force for translational deformation}. This surrogate is constructed via polynomial approximations of stiffness terms obtained from \gls{fem} data.

Based on the principle of virtual work, the joint force $\tau$ can be expressed as the derivative of the actuator's potential energy $U$ with respect to the deformation $u$, which itself depends on both the actuation parameter $p$ and the deformation $u$:
\begin{equation}\label{eq:1}
    \tau(p, u) = \frac{\mathrm{d}U(p,u)}{\mathrm{d}u} = \frac{\mathrm{\partial} U(p,u)}{\mathrm{\partial} u} + \frac{\mathrm{\partial} U(p,u)}{\mathrm{\partial} p(u)} \frac{\mathrm{d} p(u)}{\mathrm{d} u}
\end{equation}

Given the quasi-static \gls{fem} results, this derivative can be approximated using finite differences between adjacent simulation states. Distinguishing between free-loading (actuation varied, no external force) and constrained conditions (actuation fixed, external force applied), the joint forces can be decomposed into actuation and elastic contributions. By differentiating further with respect to $u$, we obtain stiffness terms $k_a(p)$ and $k_e(u)$, which depend solely on actuation and deformation, respectively. This separability allows each term to be independently approximated using low-order polynomial fits.
\textcolor{black}{Without this decomposition, the joint force would need to be approximated using a single high-order polynomial over the combined actuation–deformation space. Such models typically require significantly more data to avoid overfitting and may lead to unstable behavior when embedded within iterative physics simulations. By separating the actuation-dependent and deformation-dependent stiffness terms, the surrogate fitting problem becomes lower-dimensional and more data-efficient.}
\textcolor{black}{Such a decomposition constitutes an approximation, as it neglects higher-order coupling between actuation and deformation, and is expected to remain valid within the moderate pressure and deformation ranges considered in this study.}

The resulting surrogate provides an explicit, closed-form mapping from 
($p$, $u$) to the net force:

\begin{equation}\label{eq:2}
    \tau_n(p,u) = -\int^u_0 \big(k_a(p) + k_e(u)\big)\,\mathrm{d}u + \tau_n(p,0),
\end{equation}

where $\tau_n(p,0)$ is determined from free-loading conditions. All functions are represented by polynomial regressions fitted to \gls{fem} data, yielding a lightweight model suitable for use in rigid-body-based physics engines. Detailed derivations and an illustrative case study on a translational bellow-\gls{spa} are provided in the Supplementary Material Section~1.2 and~1.3.
\textcolor{black}{When extending the module-level surrogate to multi-module actuators through module-wise composition, additional system-level limitations arise. In particular, cumulative inter-module effects—such as compliance accumulation and boundary constraints—are not explicitly modeled, leading to a systematic underestimation of deformation as the number of modules increases, as quantified in Supplementary Figure~S2f.}

\subsection{Surrogate Model via Small Neural Networks}
\label{sec:surr_nn}
While polynomial fits offer interpretable closed-form surrogates, their approximation accuracy is inherently limited, especially when actuator behavior becomes strongly nonlinear. For example, in bellow-\gls{spa} modules with internal constraints, even higher-order fits struggle to capture the underlying relationships accurately (as demonstrated via Supplementary Figure~S3).
To address this, we employ small \gls{nn}s as alternative surrogates.

The network takes the actuation parameter $p$ and deformation $u$ as inputs and outputs the corresponding net joint force $\tau_n(p,u)$:
\begin{equation}\label{eq:3}
    (p,u) \rightarrow \tau_n (p,u)
\end{equation}
Unlike polynomial regression, \gls{nn}s flexibly approximate nonlinear relationships without requiring explicit functional assumptions. Importantly, the networks are designed to remain lightweight, ensuring efficient evaluation within \gls{prbm}-based task simulations. 

\textcolor{black}{After training, the \gls{nn}s are directly integrated into rigid-body-based physics engines such as PyBullet for task simulation, where they are used as interpolation models within the sampled operating ranges of actuation and deformation, extending applicability to more complex actuator responses.}

We implement small feed-forward \gls{nn}s trained on \gls{fem}-derived data spanning both free-loading and constrained conditions. They provide significantly improved accuracy compared with polynomial fits, while remaining lightweight for deployment in task simulations. \textcolor{black}{However, compared with the analytical polynomial surrogate, \gls{nn}-based models may produce less predictable responses when evaluated slightly outside the training distribution in physics-based simulations, and incur higher computational cost during evaluation.} Detailed architecture, training setup, and validation on a representative bellow-\gls{spa} are provided in the Supplementary Material Section~1.4.

\subsection{Meta-Model for Design-Conditioned Surrogate Generation}
\label{sec:meta_model}

The purpose of the meta-model is to establish a direct mapping from actuator design parameters \textcolor{black}{(e.g., geometric variables such as module radius, length, or thickness)} to the corresponding reduced-order surrogate representations, \textcolor{black}{enabling task-level simulations for previously unseen designs through design-space interpolation without repeating high-fidelity \gls{fem} simulations.}
By learning this design-to-behavior relationship, the meta-model enables rapid generation of design-conditioned surrogate models and therefore serves as the key interface between offline physical modeling and online task-driven design and optimization.

\begin{figure}[h]
  \centerline{\includegraphics[width=0.6\linewidth]{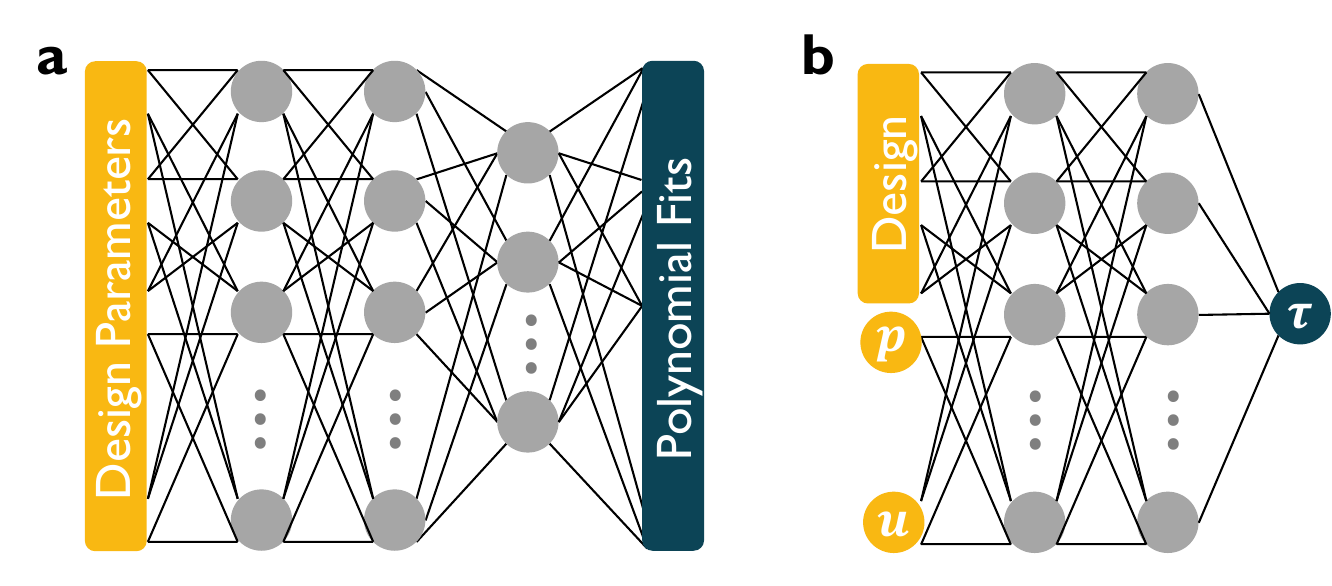}}
  \caption{Meta-models for design-conditioned surrogate generation. (a) A \gls{nn} mapping the design parameters to the corresponding polynomial coefficients of the surrogate model. (b) A \gls{nn} mapping the combined design parameters and modular actuation inputs to the resulting net force.}
\label{fig:meta_model}
\end{figure}

For surrogate models constructed using polynomial fits, the meta-model is implemented as a neural network that maps the actuator design parameters to the corresponding set of polynomial coefficients, as illustrated in Figure~\ref{fig:meta_model}(a). In this way, the polynomial surrogate for a new design can be generated instantly from its design parameters, allowing fast evaluation of actuator behavior within the simulation environment.

For surrogate models constructed using small \gls{nn}s, all available data across different designs are directly aggregated to form a unified training dataset. For each sample, the design parameters and the modular variables $(p, u)$ are concatenated as the inputs, and the corresponding net force $\tau$ is taken as the output. A \gls{nn} is then trained to capture the global relationship between the combined design–actuation inputs and the resulting behavioral response, as shown in Figure~\ref{fig:meta_model}(b). This formulation enables the surrogate to generalize jointly over both the design space and the actuation space within a single compact model.

\section{Results and Discussion}
\label{sec:results}
This section reports the performance of the proposed approach.
Two complementary evaluations are presented:
(i) the sim-to-real transfer of the surrogates on 
tasks,
and (ii) their ability to support task-driven design through optimization.
\textcolor{black}{The goal of these evaluations is to demonstrate that the proposed pipeline is not tailored to a single application, but instead offers a general modeling and optimization approach capable of supporting heterogeneous task requirements. This is addressed by applying the pipeline to tasks that differ fundamentally in both physical interaction and optimization objectives.} 
Specifically, grasping is a contact-based task in which forces are mainly transmitted through friction, and success is defined by the stability of the grasp and the ability to lift objects under load, whereas 3D shape-matching requires continuous geometric accuracy under distributed loading and gravity. 
\textcolor{black}{The following subsections present detailed, case-specific analyses for each task and actuator type. A brief general discussion at the end of this section summarizes shared assumptions and limitations across these evaluations.}
\subsection{Surrogate Models in Task Simulations}
\label{sec:sim}
This subsection focuses on the sim-to-real validation of the proposed surrogate models in \textcolor{black}{several} task simulations involving complex deformations and contact interactions. The objective is to verify that the surrogate models can accurately reproduce actuator behavior across different actuation principles and task types, without relying on full \gls{fem} simulations at runtime.

\subsubsection{Contact-Based Grasping with Polynomial Surrogate}
\label{sec:grasp}

To evaluate the capability of the surrogate models constructed using polynomial fits in task-level simulation, and to validate the proposed pipeline on a representative class of contact-dominated manipulation tasks, we simulated a soft gripper performing grasping, \textcolor{black}{where the smooth and efficient numerical characteristics of polynomial surrogates facilitate stable physics-based simulation}. The gripper design is based on the default bellow-\gls{spa} from prior work, featuring a strain-limiting side that constrains expansion on one face and thereby promotes directional bending during pressurization~\cite{Yao2022}.

\begin{figure}[h]
\centerline{\includegraphics[width=0.7\linewidth]{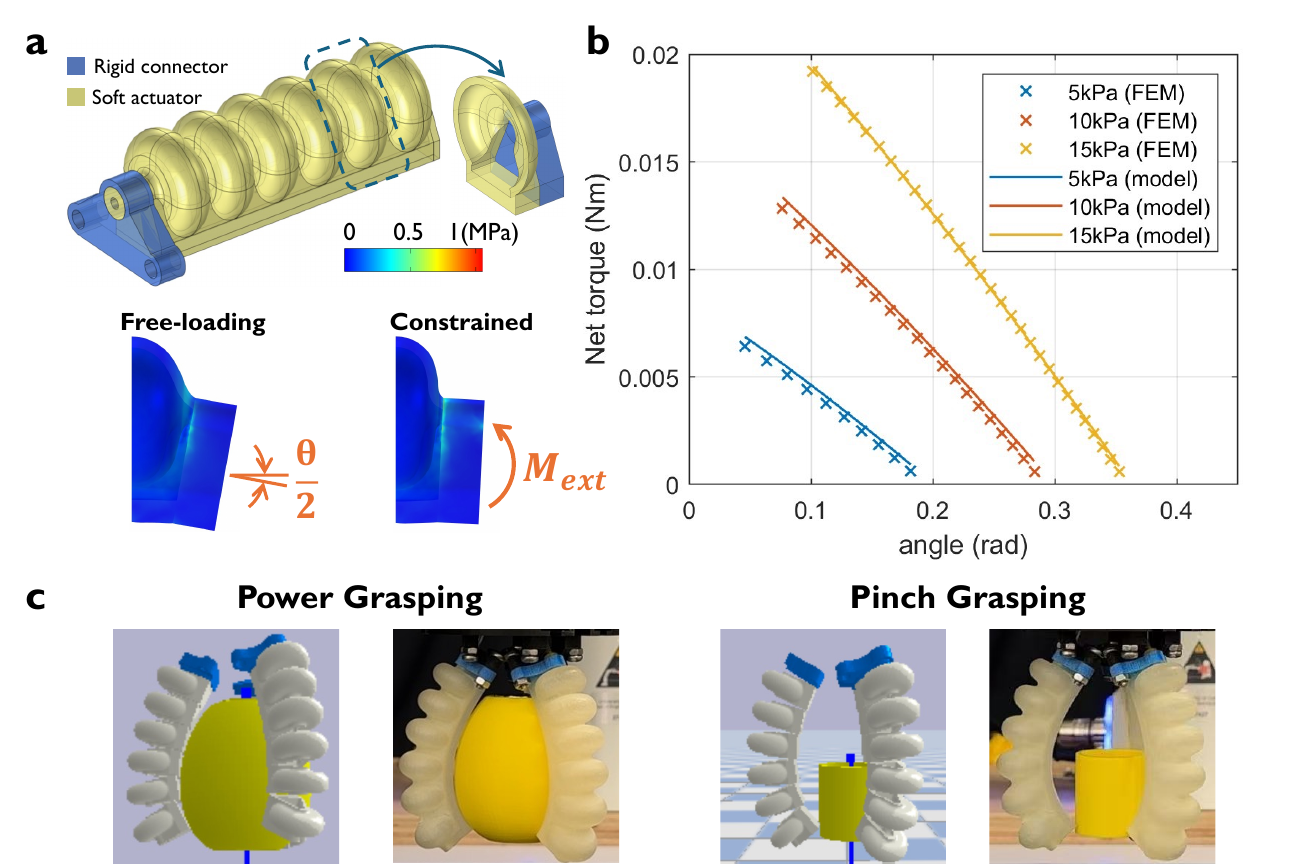}}
  \caption{Validation of the polynomial-based surrogate model through a soft gripper grasping task. (a) \gls{fem} model of the bellow-\gls{spa} used in the gripper. (b) Comparison of net torque vs. angular deflection under different pressure levels obtained from the \gls{fem} simulation and the surrogate model. (c) Two grasping modes with different objects, compared between simulation and experiment.}
\label{fig:grasp}
\end{figure}

As illustrated in Figure~\ref{fig:grasp}a, we simulated a half-module, closed with an end cap to facilitate smooth deformation, under conditions of free pressurization and externally applied moment $M_{ext}$ at various fixed internal pressures, obtaining angular deflection $\theta$. \textcolor{black}{In this study, the pressure is sampled from $0$~kPa to $15$~kPa with $0.5$~kPa as an interval, and the externally applied moment is sampled from $0~\mathrm{N\cdot m}$ to $0.02~\mathrm{N\cdot m}$ with an interval of $0.0006~\mathrm{N\cdot m}$.}
Following the previously described modeling process, we compared the model-predicted net torque to the applied external moment across different pressure values, resulting in a maximum error of 8.26\% relative to the moment range.

\textcolor{black}{Grasping simulations were then carried out in PyBullet using a physics-based model of the soft gripper. The gripper consists of three actuator fingers equally spaced around a central axis, with each finger modeled as a chain of links connected by revolute joints. \textcolor{black}{Visual and collision geometry are taken directly from the actuator CAD STL mesh files, while the mass properties of each link are estimated from the actuator volume and material density.} The lateral friction coefficient between the actuator and PLA-printed objects is set to 0.8~\cite{FrictionRubber2012}, and anchor friction is enabled to mimic soft contact behavior. To better approximate real-world conditions, 5\% random noise is introduced to the object placement, friction coefficient, and actuation pressure in the simulations. To ensure a fair sim-to-real comparison, the simulated grasping setup is configured to match the physical experiments in terms of gripper geometry, object placement, and actuation conditions. Details of the ablation study used for selecting the simulation parameters, together with the corresponding single-run simulation time, are reported in the Supplementary Material Section~2.1.}

\textcolor{black}{Using this simulation setup, we evaluated sim-to-real transfer by comparing grasping performance against physical experiments.}
Two representative objects were considered: a 3D-printed egg-shaped object (maximum radius: 30 mm; height: 65 mm) for power grasping, and a cylindrical object (radius: 16 mm; height: 35 mm) for pinch grasping. The objects weighed 50 g and 10 g, respectively. By systematically increasing the object mass by adding standardized 20~g weights \textcolor{black}{(at a predefined central location integrated into the object design prior to fabrication)} and applying different actuation pressures, we evaluated how closely the simulated grasping success rates align with experimental outcomes. Specifically, grasping was conducted under two actuation pressures, 15~kPa and 13~kPa. For each grasping mode, three weight levels were tested: for power grasping of the egg-shaped object, 7, 6, and 5 units of 20~g were added, while for pinch grasping of the cylinder, 4, 3, and 2 units of 20~g were used. \textcolor{black}{Despite the small sample sizes and the use of binary and coarse success metrics, this sim-to-real validation is intended to demonstrate the feasibility of transferring surrogate-based task-level simulation outcomes to physical grasping behavior, rather than to provide a comprehensive performance benchmark.}

\begin{figure} [h]
\includegraphics[width=\linewidth]{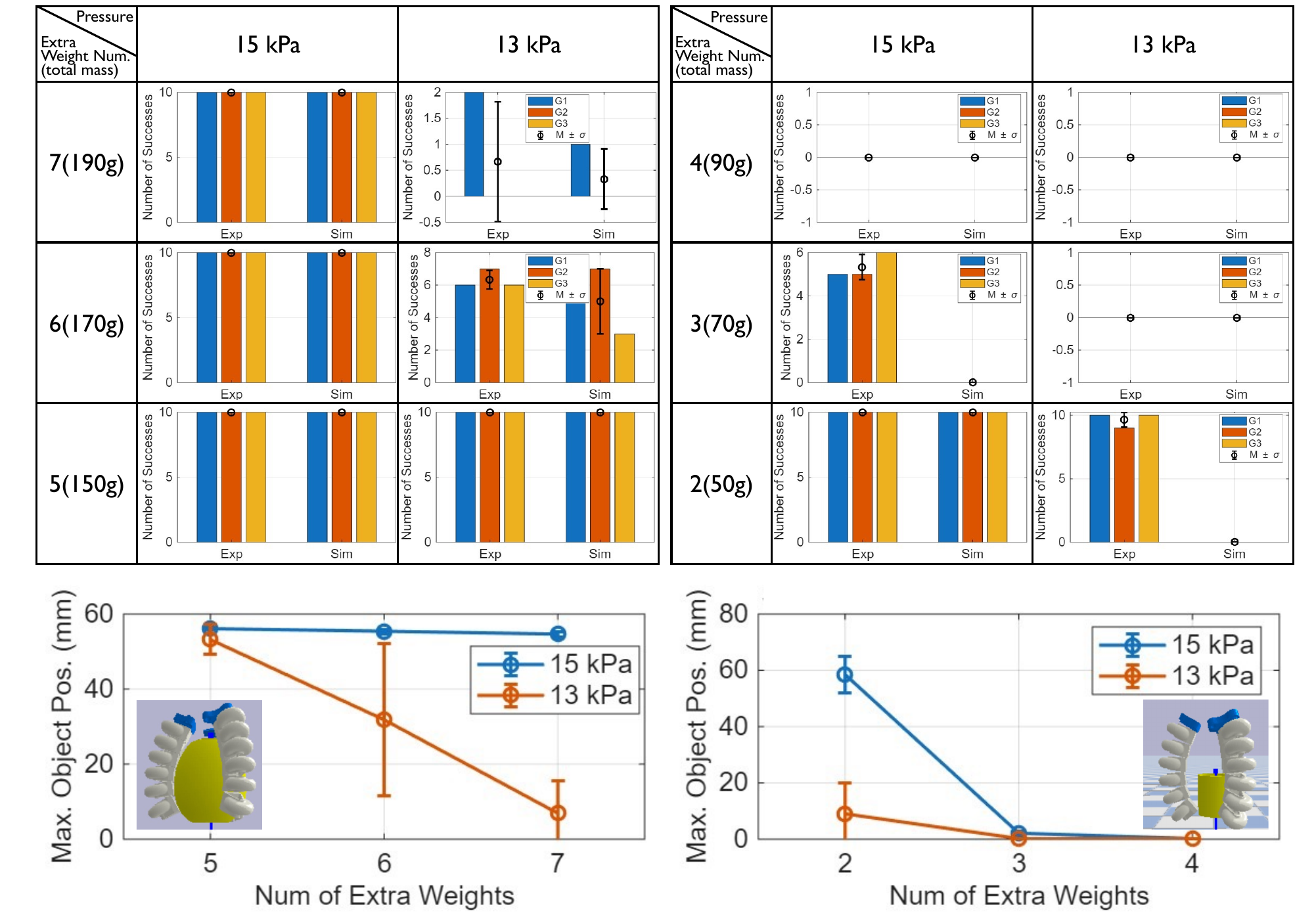}
  \caption{Comparison of experimental and simulated grasping results for the egg (left) and the cylinder (right). The upper tables report the number of successful grasps across different actuation pressure levels and added weight conditions. \textcolor{black}{G1, G2, and G3 denote three independent experimental sets, each consisting of 10 grasping trials conducted under identical conditions. The lower plots show the maximum vertical ($z$-axis) position of the simulated object relative to the initial placement ($z=0$)}, as a function of the applied weight under different actuation pressures.}
\label{fig:grasp_results}
\end{figure}

Physical experiments were conducted using a Franka Emika\textsuperscript{\textregistered} Panda arm (Franka Robotics, Germany) with the soft gripper mounted as the end effector and controlled through its standard interface. The actuation pressure was supplied by an Anest Iwata\textsuperscript{\textregistered} air compressor (Anest Iwata Corporation, Japan) and regulated by a VPPE-3-1-1/8-2-010-E1 Festo\textsuperscript{\textregistered} proportional-pressure regulator (Festo AG \& Co. KG, Germany). At each combination of weight level and pressure setting, three sets of ten grasping trials were performed, yielding three success rates (out of 10) for each object under each condition. \textcolor{black}{During each trial, the pneumatic actuation was applied over approximately 2–3 seconds for the grasp attempt, allowing the actuator to establish contact with the object and apply grasping force.} Representative examples from both simulations and experiments are provided in the Supplementary Video.

The success rate comparisons are shown in the upper tables of Figure~\ref{fig:grasp_results}. Overall, the simulation tends to underestimate the actual grasping performance, which is expected as certain inter-modular effects are not captured by the model. Nevertheless, the simulated trends are in good agreement with the experimental results.

Additionally, the lower plots in Figure~\ref{fig:grasp_results} show that, although the success rates are underestimated in simulation, the simulated maximum $z$-position $z_{\max}$ of the object correlates with the experimental performance. A higher simulated $z_{\max}$ is associated with a higher success rate in physical experiments, suggesting it can serve as a useful performance indicator.

\subsubsection{Continuous Deformation with Small NN Surrogates}
\label{sec:helix}

To evaluate the capability of the surrogate models constructed with small \gls{nn}s in task-level simulation, and to further validate the proposed pipeline on a representative class of continuous deformation tasks distinct from contact-dominated grasping, we simulated an elephant-trunk-inspired helical soft actuator, adapted from prior work and incorporating a central thread to constrain elongation~\cite{Yao2024}, under gravity and externally applied tip loads.

\begin{figure}
  \centerline{\includegraphics[width=0.9\linewidth]{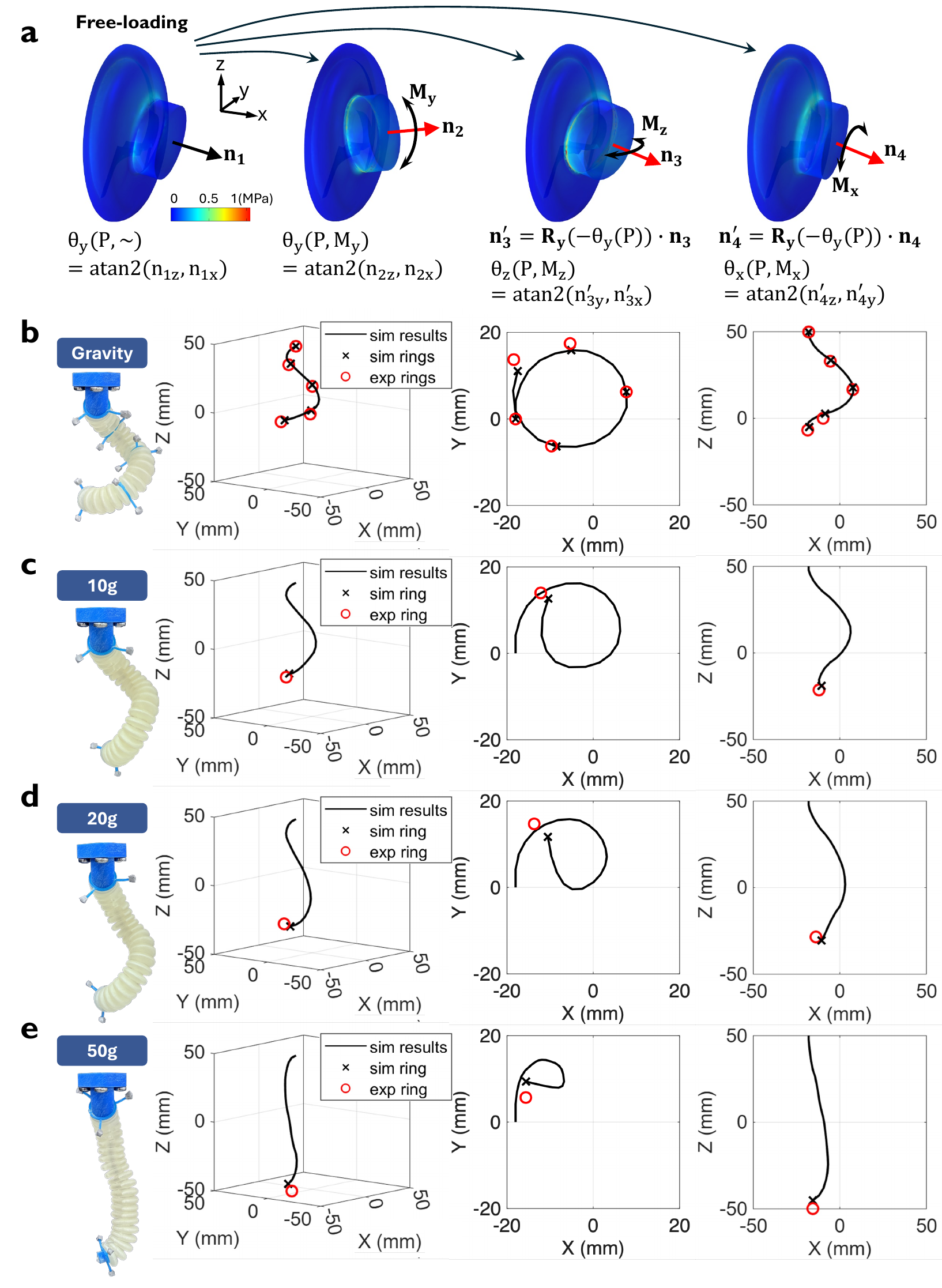}}
  \caption{Validation of the surrogate model based on small \gls{nn}s using a helical actuator under gravity and various tip loads. (a) \gls{fem} model of the bellow-\gls{spa} used in the helical actuator for all three axes. (b) Comparison of the actuator deformation between simulation and averaged physical experiments over three trials under gravity at five ring positions. (c–e) Comparison of the actuator deformation between simulation and averaged physical experiments over three trials under tip loads of 10, 20, and 50~g applied at the actuator tip.}
\label{fig:helix_results}
\end{figure}

As shown in Figure~\ref{fig:helix_results}a, \gls{fem} simulations were performed to characterize the deformation of a half-module under two conditions: (i) free-loading pressurization with increasing internal pressure, and (ii) constant pressure combined with externally applied moments about each principal axis. When moments were applied around the non-active axes (z and x), the rotation about the y-axis was first reversed to isolate angular deflections $\theta_z$ and $\theta_x$. \textcolor{black}{In this case, the internal pressure is sampled from $0~\mathrm{kPa}$ to $15~\mathrm{kPa}$ with an interval of $0.5~\mathrm{kPa}$. The externally applied moments are sampled independently along each axis, with $M_y$ ranging from $0$ to $0.8~\mathrm{N\cdot m}$ at intervals of $0.05~\mathrm{N\cdot m}$, and $M_x$ and $M_z$ ranging from $0$ to $0.006~\mathrm{N\cdot m}$ at intervals of $0.0002~\mathrm{N\cdot m}$.}
\textcolor{black}{Following this axis-wise loading and deformation extraction procedure,}
three \gls{nn}s were trained to capture the mappings between pressure, moment, and angular displacement about each axis, each achieving $R^2 > 0.999$.
\textcolor{black}{This axis-wise modeling choice constitutes a simplification that reduces learning complexity and improves training stability. Cross-axis coupling effects may become more pronounced under larger deformations or combined multi-axis loading.}

To validate the surrogate model, we simulated actuator deformation in PyBullet and compared the results with real-world experiments under self-weight (Figure~\ref{fig:helix_results}b) and additional tip loads of 10g, 20g, and 50g (Figure~\ref{fig:helix_results}c-e). In PyBullet, the actuator was modeled as a chain of links connected by spherical joints, with geometry and mass properties matched to the corresponding \gls{spa} modules. The tip load was simulated as a downward external force applied to the terminal link. Details of the ablation study used for selecting the helical actuator simulation parameters, as well as the associated computational cost per simulation run, are provided in the Supplementary Material Section~2.2. Recordings of the simulations are provided in the Supplementary Video.

Experimental deformation data were recorded using a Motive OptiTrack\textsuperscript{\texttrademark} system (NaturalPoint, Inc., USA) with four Flex 3 cameras, tracking the positions of five rings (each comprising three reflective markers) along the actuator, following the procedure in~\cite{Yao2024}. The actuation pressure was set to 15~kPa and supplied using the same pneumatic setup as in Section~\ref{sec:grasp}.
\textcolor{black}{This pressure level was empirically identified as the maximum representative operating pressure for this actuator. As the surrogate is trained on FEM data spanning $0$–$15~\mathrm{kPa}$, its predictions are expected to remain accurate within this range as an interpolation problem, with validation performed at the upper operating limit.}

Each experiment was repeated three times to compute average ring positions. \textcolor{black}{To ensure quasi-static behavior during these measurements, we waited approximately 20–30 seconds after each actuation step before recording the actuator configuration.} For comparison, the actuator base position in PyBullet was aligned with the experimentally measured mean position of the base ring. Under self-weight, the relative errors at successive ring positions were 1.52\%, 1.08\%, 2.65\%, and 2.85\% of the actuator length, with \textcolor{black}{a maximum experimental standard deviation of 1.29 mm observed at the end ring.} Under additional tip loads, the relative tip errors were 2.97\%, 4.25\%, and 5.31\% for 10g, 20g, and 50g loads, respectively, with \textcolor{black}{a maximum experimental standard deviation of 1.33 mm at 20g.}

\subsubsection{A Tendon-Driven Finger Simulation}
\textcolor{black}{To demonstrate the applicability of the proposed modeling approach beyond bellow actuators and pneumatic actuation,}
we also applied the \gls{nn} surrogate model to a tendon-driven actuator from prior work~\cite{Capp2025}, simulating its deformation under free loading and tip loads of 20g and 50g with the same tendon travel.

\begin{figure}[h]
  \centerline{\includegraphics[width=0.75\linewidth]{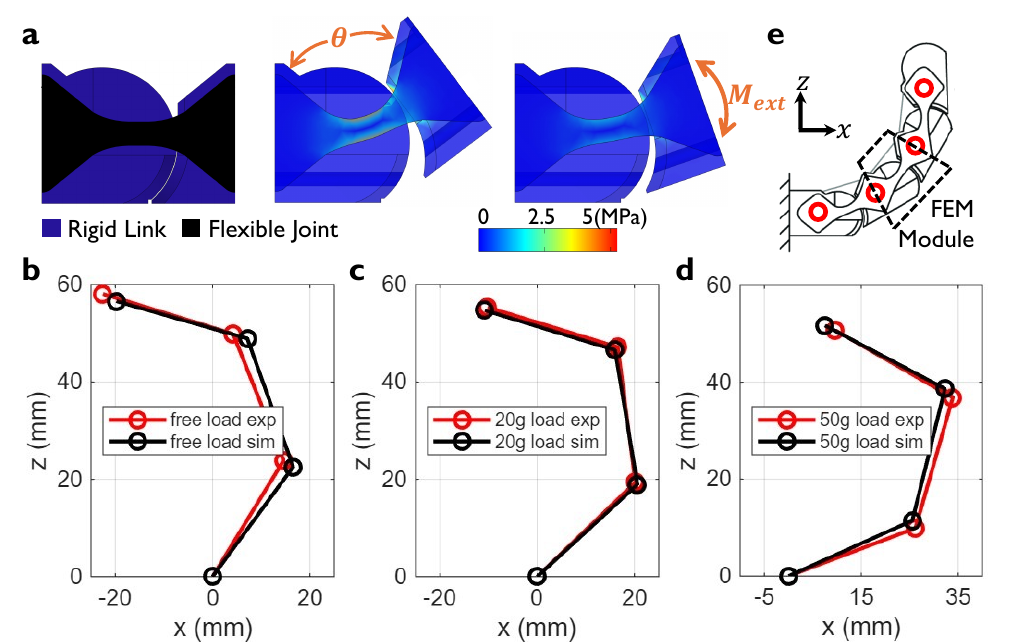}}
  \caption{Validation of the surrogate modeling approach beyond the bellow-\gls{spa}. (a) \gls{fem} model of the joint in the tendon-driven finger developed by~\cite{Capp2025}. (b–d) Comparison of link positions between simulation and averaged experimental results under free load, 20~g, and 50~g tip loads, respectively. \textcolor{black}{Circle markers correspond to the center positions of each rigid link, with the base link center used as the origin of the reference frame, as indicated in (e), adapted with permission from~\cite{Capp2025}. Copyright \textcopyright{} 2025 IEEE.}}
\label{fig:tendon_results}
\end{figure}

As shown in Figure~\ref{fig:tendon_results}a, \gls{fem} simulations were conducted to characterize the deformation $\theta$ of an actuator module (as indicated by the dashed box in Figure~\ref{fig:tendon_results}e) under free loading and externally applied moments $M_{ext}$, with the tendon force modeled as a directional load following the tendon path. \textcolor{black}{In this case, the tendon force is sampled from $0~\mathrm{N}$ to $8~\mathrm{N}$ with an interval of $0.05~\mathrm{N}$. The externally applied moment is sampled from $0$ to $0.05~\mathrm{N\cdot m}$ at intervals of $0.0025~\mathrm{N\cdot m}$.}
A \gls{nn} was trained to capture the mapping between tendon force, external moment, and angular deflection, achieving $R^2 > 0.999$. In addition, a simple regression was used to map tendon force to tendon travel.

To validate the sim-to-real transfer, the tendon-driven actuator was modeled in PyBullet as a chain of links connected by revolute joints, with geometry and mass properties taken from the CAD model. The tip load was simulated as an external force applied to the terminal link and directed toward the pulley position used in the real experiments, which corresponds to the finger’s initial tip position. Simulation results were obtained by fine-tuning the tendon force to reproduce the tendon travel reported in~\cite{Capp2025}. The PyBullet simulation parameters for this tendon-driven actuator were chosen to be identical to those used in the helical actuator experiments, as no object contact was involved. Recordings of the simulations are provided in the Supplementary Video.

\textcolor{black}{The simulation results show good agreement with experimental measurements reported in~\cite{Capp2025}, where each loading condition was experimentally repeated three times, with a maximum standard deviation of 3.51 mm at the end link position under free load.} As illustrated in Figure~\ref{fig:tendon_results}b-d, the resulting tip errors were 3.14 mm under free loading, 0.84 mm under a 20 g tip load, and 2.29 mm under a 50 g tip load.

\subsection{Surrogate Models for Task-Driven Designs}
\label{sec:opt}
In this section, we evaluated the effectiveness of the proposed modeling pipeline for task-driven design by coupling the meta-models with \gls{sota} optimization algorithms. Two representative design studies are considered: (i) soft gripper co-design using reinforcement learning, and (ii) 3D actuator shape matching using evolutionary optimization. These two cases employ different meta-model instantiations and optimization strategies, thereby demonstrating that the proposed pipeline can flexibly incorporate task-specific surrogate representations and remain fully compatible with modern learning-based and gradient-free optimization methods.

\subsubsection{Designing Soft Gripper with RL}
\label{sec:RL}

\begin{figure}[t]  \includegraphics[width=\linewidth]{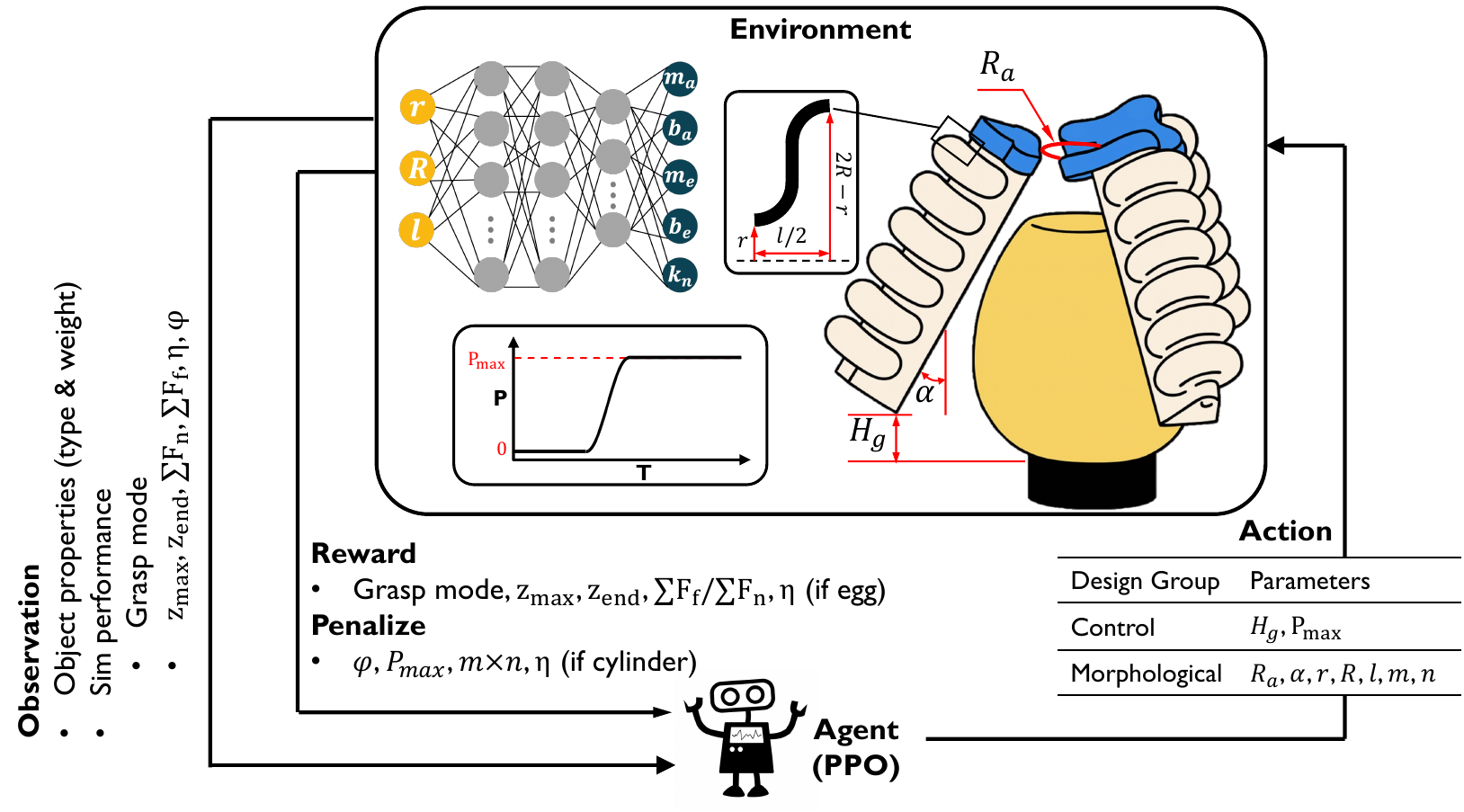}
  \caption{Schematic overview of task-driven soft gripper design using \gls{rl}. An \gls{rl} agent (PPO) is trained on a task distribution consisting of grasping an egg and a cylinder with varying weights. The environment corresponds to a simulated grasping trial with randomized object type and weight, where both control and morphological parameters are sampled from the action space and evaluated using the meta-model for polynomial surrogates. The agent observes task properties and simulated grasping performance features, and assigns rewards and penalties accordingly to learn a policy that maps object-specific conditions to optimal gripper design parameters.}
\label{fig:RL}
\end{figure}

The first case study employs the polynomial-based meta-model in a \gls{rl}-driven soft gripper co-design setting, as illustrated in Figure~\ref{fig:RL}. \textcolor{black}{Despite its simplified structure, the polynomial surrogate has been shown to provide sufficient accuracy and smoothness for the considered bellow-\gls{spa}-based soft gripper within the investigated operating range (Section~\ref{sec:grasp}), while maintaining the computational efficiency required for RL-based design exploration.} Here, both morphological design parameters and actuation-related variables are optimized jointly through task-level grasping performance, with the meta-model providing fast predictions of module-level mechanical responses during simulation.

We first constructed the meta-model from a family of polynomial-based surrogates. The same polynomial surrogate modeling procedure as described in Section~\ref{sec:grasp} was applied repeatedly over a discretized design space defined by the actuator inner radius $r \in \{2.75, 3.5, 4.25, 5\}$~mm, wall thickness $t = 1.5$~mm, average radius $R \in \{6.75, 7.5, 8.25, 9\}$~mm, and module length $l \in \{9, 10, 11, 12\}$~mm. \textcolor{black}{This design space is chosen as a local neighborhood around the baseline gripper design used in Section~\ref{sec:grasp}, enabling task-driven optimization within a practically relevant actuator configuration range} \textcolor{black}{while avoiding unrealistic actuator geometries that would violate the manufacturability and geometric constraints of the bellow-\gls{spa} design.} This resulted in 54 valid actuator designs, with the remaining combinations excluded by the geometric boundary conditions imposed on the design parameters~\cite{Yao2022}.

A meta-model was then trained as described in Section~\ref{sec:meta_model}, using a three-dimensional input vector composed of the design parameters ($r, R, l$). The network comprises three hidden layers, two with 64 neurons and one with 32 neurons, each using ReLU activation, followed by a linear output layer predicting the five polynomial coefficients ($m_a$, $b_a$, $m_e$, $b_e$, $k_n$), \textcolor{black}{as defined in the polynomial surrogate formulation detailed in the Supplementary Material Section~1.3, Equation~S7).} The model was trained with the Adam optimizer to minimize the \gls{mse} for up to 1000 iterations and achieves an $R^2$-score exceeding 0.99 on the full dataset.

This meta-model was embedded into the PyBullet simulation to replace explicit per-design polynomial surrogates, enabling prediction of the actuator’s grasping response for continuously varying design parameters. 
\textcolor{black}{The actuator geometries corresponding to different design parameters are generated by scaling the baseline STL mesh used in Section~\ref{sec:grasp}.}
The task-level simulation takes three groups of inputs. The control domain contains the grasping height $H_g$ (distance between the actuator tip and the object bottom) and the maximum pressure $P_\mathrm{max}$.
The morphological domain includes the actuator base radius $R_a$ (radius of the circle formed by the actuator bases), inclination angle $\alpha$ (angle between the actuator longitudinal axis and gravity), the geometric parameters ($r, R, l$), the module number $n$ (modules per actuator) and the actuator number $m$ (number of actuators).
The object properties comprise the object type (“egg" or “cylinder") and the weight number (the number of attached 20 g weights). All other simulation settings follow those in Section~\ref{sec:grasp}.

The simulator outputs a set of task-relevant features: the maximum and final object positions along the vertical axis $z_{\max}$, $z_{end}$; the summed normal and friction contact forces over the grasping stage $\sum F_n$, $\sum F_f$; the contact module efficiency $\eta$ (ratio between the number of active contact modules at the end of grasping and the total module count); the slip ratio $\varphi$ (ratio between the average object velocity and gripper velocity along the $z$-axis); and the grasping mode (miss, touch, slip, or stable grasp) determined from object motion and contact forces.

On top of this simulation, we constructed a \gls{rl} environment for task-driven gripper design. The action space contains all control and morphological design variables listed above, while the observation space includes the object properties together with the simulated grasping performance features. The reward function encourages successful grasping modes, higher lifting height, and larger friction-to-normal force ratios, while penalizing excessive pressure and material usage. It further promotes distributed multi-module contacts when grasping the egg and more localized contacts when grasping the cylinder.

Task-driven design was then performed using a \gls{ppo} agent trained on a task distribution consisting of grasping an egg and a cylinder with varying weights. Each episode corresponds to a single simulated grasp attempt, with the object type and weight randomized at reset, so that the learned policy maps task-dependent object conditions to both control commands and gripper morphology. Training was run for 14400 total timesteps in a vectorized environment with four parallel simulations, using the hyperparameters detailed in Supplementary Material Section~2.3. The associated learning curves, shown in Figure~S8, indicate stable convergence of the policy and value function.

The resulting distributions of the optimized parameters are summarized in Figure~S6 and~S7, obtained by evaluating the trained policy on each object type--weight combination and sampling the predicted action parameters over 10 stochastic rollouts per task condition. Most morphological design variables, including inner radius, average radius, module length, module number, and actuator number, converge to nearly identical values across all tasks, whereas the actuator base radius exhibits a clear task-dependent differentiation between the egg and cylinder. This suggests that, under the current reward and task distribution, the policy predominantly converges to a robust general-purpose gripper morphology, with a small subset of parameters adapting to object geometry.

\begin{figure}[h]
\centerline{\includegraphics[width=0.6\linewidth]{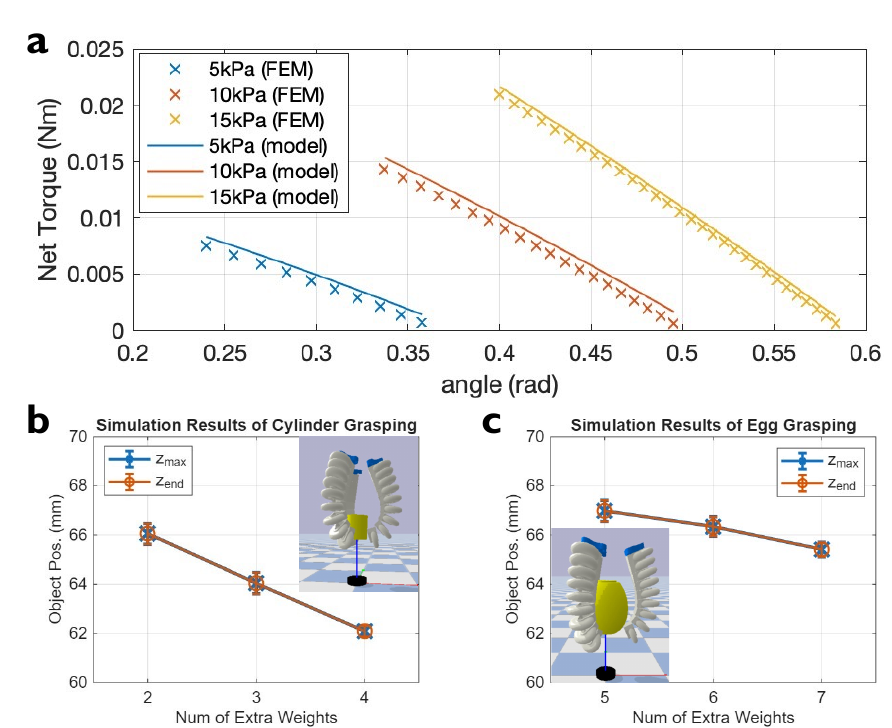}}
  \caption{Results of the \gls{rl}-optimized design. (a) Comparison of net torque versus angular deflection under representative pressure levels between \textcolor{black}{the polynomial surrogate model for the optimized design and its \gls{fem} simulation, showing close agreement consistent with the validation in Section~\ref{sec:grasp}}. (b) Simulated object grasping results using the polynomial surrogate model for the optimized design, indicating successful and stable grasping under the intended task conditions.}
\label{fig:RL_resim}
\end{figure}

To further verify the quality of the learned design and to remove potential uncertainty introduced by the meta-model, we re-computed the polynomial surrogate for the final optimized design parameters using direct \gls{fem} data. Notably, this design does not belong to the original meta-model training set. The reconstructed surrogate achieved a \gls{rmse} of 3.21\% relative to the maximum net torque value when compared with \gls{fem} data (Figure~\ref{fig:RL_resim}a). Using this refined surrogate, grasping simulations (10 attempts with added environmental noise) were carried out for all task conditions. 
\textcolor{black}{
The simulation protocol follows the same procedure used in Section~\ref{sec:grasp} for the sim-to-real validation of the baseline gripper. 
As shown in Figure~\ref{fig:RL_resim}b and c, the simulated maximum z-position $z_{\max}$ and final z-position $z_{\text{end}}$ of the object are nearly identical when grasped by the optimized gripper, indicating stable grasps and corresponding to a success rate of 10/10 in all trials.
A quantitative comparison with the baseline gripper configuration reported in Section~\ref{sec:grasp} is provided in the Supplementary Table~S1, which summarizes the grasp success rates and the simulated maximum object position $z_{\max}$ across the tested task conditions.}
\textcolor{black}{These results confirm that the proposed pipeline can exploit \gls{rl} together with the polynomial-based meta-model to support task-driven soft gripper design optimization.}

\subsubsection{3D Shape-Matching with CMA-ES}\label{sec:sm}

\begin{figure}[h]
  \centerline{\includegraphics[width=0.9\linewidth]{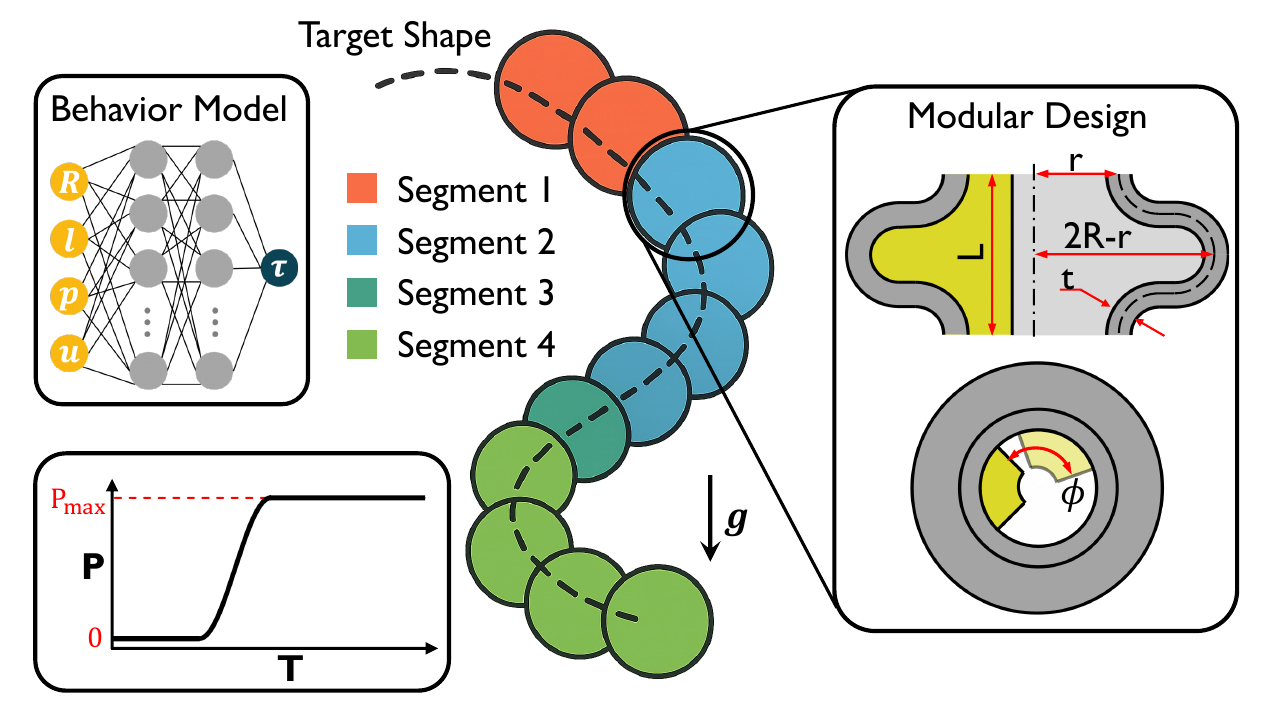}}
  \caption{Schematic overview of the shape-matching actuator design task. The actuator is composed of multiple segments (highlighted in different colors), each characterized by the number of identical modules, the geometric design parameters of the module, and the rotation angle relative to the preceding segment. Behavior models that map the modular design and actuation inputs to the output response are employed in simulation to obtain the deformed shape of the designed actuator under actuation pressure and gravity, in order to match a given target shape.}
\label{fig:shape-matching}
\end{figure}

The second case study uses the \gls{nn}-based meta-model within a representative evolutionary optimization method for actuator shape-matching design, as shown in Figure~\ref{fig:shape-matching}. The meta-model predicts the deformation behavior of candidate designs, while a gradient-free optimizer updates the design parameters according to a geometric matching objective. \textcolor{black}{In this case study, the shape-matching task is defined using the same right-handed helical target shape as in~\cite{Yao2024}, with a radius of 18 mm and a height of 60 mm.}

We first applied the \gls{nn} surrogate model to the same actuator design for deformation prediction along the three principal axes as described in Section~\ref{sec:helix}. This procedure was repeated over a discretized design space defined by the average radius
$R \in \{5, 5.5, 6, 6.5, 7\}$~mm and module length
$l \in \{4, 4.5, 5, 5.5, 6\}$~mm, while keeping the inner radius ($r = 3$~mm) and wall thickness ($t = 1.5$~mm) fixed. \textcolor{black}{This design space is centered around the optimal actuator configuration reported in~\cite{Yao2024}, allowing interpolation-based evaluation of the meta-model in the vicinity of a known high-performance design.} After excluding combinations that violate the geometric condition $R - r \geq l/4$~\cite{Yao2024}, 24 valid actuator designs remained.

Following Section~\ref{sec:meta_model}, a behavior model was trained for each axis by aggregating data from all sampled designs. For a design characterized by $(R, l)$, all tuples ($p, u, \tau$), corresponding to actuation pressure, half joint angle, and resulting joint torque, were collected and stacked into input–output pairs of the form $(R, l, p, u) \rightarrow \tau$. Each axis-wise behavior model was implemented as a feedforward NN with two hidden layers of 64 neurons and ReLU activation, followed by a linear output layer predicting the net joint torque. The models were trained with the Adam optimizer to minimize \gls{mse}, achieving $R^2$ scores exceeding 0.999 for all three axes.

These design-conditioned behavior models were then embedded into the PyBullet simulation environment to replace the explicit small \gls{nn} models, enabling fast prediction of actuator deformation for varying design parameters. The simulation takes two groups of inputs. The first defines the actuator design: the number of segments $N$; the average radius $R$ and module length $l$ shared by all modules within each segment; the number of modules per segment $n$; the relative rotation angle between adjacent segments $\phi$, and the actuation pressure $P_{\max}$. The second group specifies the target shape to be matched.

\begin{figure}[h]
  \includegraphics[width=\linewidth]{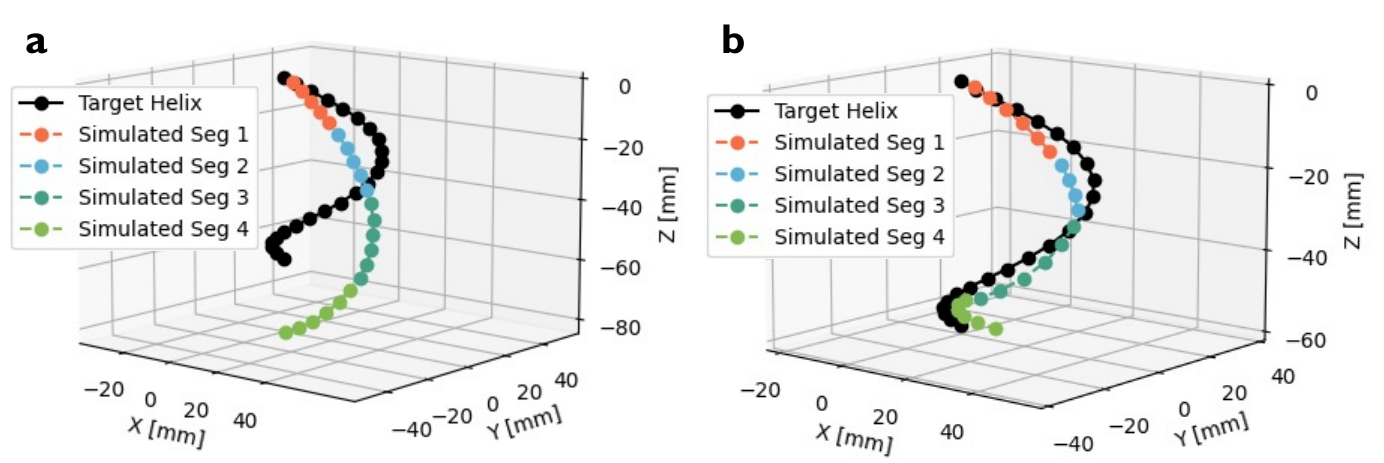}
  \caption{Results of the shape-matching design optimization. (a) Comparison between the target helix and the optimized actuator obtained using the previous modeling approach~\cite{Yao2024}. (b) Comparison between the target helix and the optimized actuator obtained using the proposed behavior models.}
\label{fig:sm}
\end{figure}

For each candidate design and pressure, the PyBullet simulation produces the deformed centerline of the actuator, reconstructed as a piecewise arc-length-parameterized curve based on link positions and joint rotations. The target helix is re-sampled according to the normalized cumulative arc length of this curve to ensure one-to-one correspondence between simulated and target points. Shape-matching quality is quantified using the \gls{rmse} and the maximum point-wise Euclidean distance $e_{\max}$ between corresponding points, and these metrics serve as objective functions for design optimization. In~\cite{Yao2024}, no satisfactory actuator could be identified due to limitations of the underlying model, with the best design yielding an \gls{rmse} of $25.68~\mathrm{mm}$ and an $e_{\max}$ of $36.20~\mathrm{mm}$ (Figure~\ref{fig:sm}a).

The optimization problem is formulated as a joint search over structural parameters, geometric design parameters, and actuation pressure. At the structural level, the actuator is partitioned into $N=4$ segments in this study. The total number of modules is first bounded using a physically motivated estimate based on the target curve length and the range of module lengths, and then five candidate totals are sampled around the center of this interval. For each total, multiple structural configurations ${n_i}_{i=1}^{N}$ are generated by distributing modules across segments under a balance constraint, avoiding extreme structural imbalance while preserving diversity. This procedure yields 81 structural candidates.

For each structural candidate, the continuous design parameters $(R_i, l_i, \phi_i)_{i=1}^{N}$ and actuation pressure $p$ are optimized jointly. To remove global rotational redundancy, the first inter-segment rotation is fixed to $\phi_1 = 0$, while the remaining $\phi_i$ are treated as optimization variables. Continuous optimization is carried out using the \gls{cma-es}, which is well suited to strongly nonlinear, non-convex objective landscapes and black-box settings where each function evaluation involves a full physics-based simulation with embedded surrogate models. The initial step size is set to $\sigma_0 = 0.3$ to balance global exploration and local refinement.

During the shape-matching process, \gls{cma-es} iteratively searches the continuous design space for each of the 81 structural configurations, minimizing both \gls{rmse} and $e_{\max}$ until either convergence below 5 mm for both metrics or 60 iterations are reached. The resulting optimal actuator design and pressure are reported in Supplementary Material Section~2.4, and summarized in Table~S2. Based on the behavior model prediction, the optimized design achieved an \gls{rmse} of $4.51~\mathrm{mm}$ and an $e_{\max}$ of $8.46~\mathrm{mm}$ with respect to the target helix.

To further evaluate the fidelity of the behavior model prediction, the small \gls{nn} surrogate models for each segment were subsequently reconstructed using direct \gls{fem} data. All reconstructed surrogates achieved $R^2$ scores above 0.999. The corresponding actuator was then simulated again, yielding an \gls{rmse} of $ 5.05~mm$ and $e_{\max} = 8.19~mm$ with respect to the target helix (Figure~\ref{fig:sm}b). Compared with the previous best result, this represents a substantial improvement in shape matching and confirms that the \gls{nn}-based behavior models, when combined with a representative evolutionary optimizer such as \gls{cma-es}, are effective for task-driven design within the proposed pipeline.

\subsection{General Discussion and Limitations}
Building on the results presented above, this subsection provides a brief general discussion of the limitations underlying the proposed approach.

\textcolor{black}{First, the proposed pipeline primarily supports interpolation within parameterized actuator design spaces that are explicitly sampled during surrogate construction, and does not aim to generalize across fundamentally different morphologies or actuation principles. Second, the surrogate models were constructed from quasi-static FEM data, which neglect dynamic effects such as rate-dependent material behavior and transient interactions. Third, while extensive validation was performed mainly within the sampled ranges, systematic evaluation outside these ranges remains limited. Finally, hardware validation was demonstrated for selected representative cases, but not for all actuator types and task scenarios considered in simulation.}

\section{Conclusion}
This paper introduced a unified FEM-based surrogate modeling pipeline for generalized task-driven design of soft robots. By integrating high-fidelity \gls{fem} data generation, compact surrogate modeling, design-conditioned meta-model learning, and \gls{prbm}-based task-level simulation, the proposed pipeline bridges the long-standing gap between physical accuracy and computational efficiency in soft robotic design. Through extensive sim-to-real validation across multiple actuator types and two representative task-driven design studies, namely \gls{rl}-based soft gripper co-design and evolutionary shape-matching optimization, we demonstrated that the pipeline supports both reliable physical prediction and efficient task-level optimization under realistic environmental interactions.
\textcolor{black}{Beyond the specific tasks investigated in this work, the modular structure of the proposed pipeline facilitates extension to additional soft robotic applications within similar modeling assumptions.}
In particular, the reliance on physically grounded \gls{fem} data and surrogate abstraction makes the pipeline well suited for environments involving complex interactions, such as underwater manipulation, fluid–structure coupling, and confined-space operations. Moreover, the decoupling between surrogate modeling and task-level optimization allows the proposed approach to be readily combined with diverse learning-based and gradient-free optimization paradigms. These properties position the pipeline as a fundamental building block toward scalable, autonomous, and general-purpose task-driven design of soft robotic systems in complex real-world environments.




\section*{Supplementary material}

# Supplementary Material

## 1 Material and Methods

This section complements Section 3 of the main text by providing the detailed derivations and implementation details. It covers the FEM setup, surrogate model derivations, and illustrative examples, and also provides additional results for polynomial and neural network (NN) surrogates.

### 1.1 FEM Setup for Data Generation

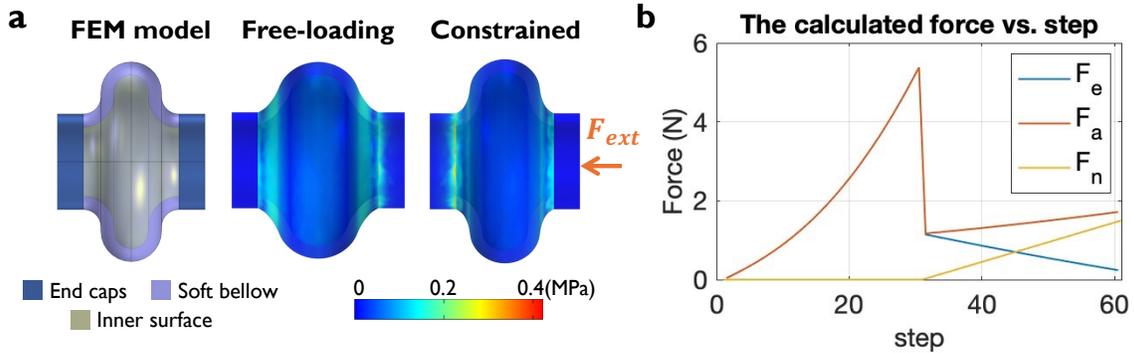

Figure S1: FEM setup and force evolution for a translational bellow-SPA module.(a) The FEM model under free-loading and constrained conditions.(b) Calculated elastic, actuation, and external forces over the simulation steps, illustrating the transition from the free-loading phase to the constrained phase.

To illustrate the FEM procedure described in Section 3.1, we used a modular translational bellow-SPA as an example. The actuator is fabricated with Stratasys digital materials and features a U-shaped bellow geometry driven by pneumatic actuation. All soft pneumatic actuators considered in this work (except for the tendon-driven actuator) are modeled or printed using Agilus30™elastomer (Stratasys Ltd.), and its hyperelastic behavior in the FEM simulations is defined based on the stress–strain characterization reported in [1].

A stationary 3D FEM model, shown in Figure S1a, was constructed in COMSOL Multiphysics® (COMSOL Inc., Sweden) to approximate quasi-static conditions, which are appropriate for low-speed operation. The physical model is based on a prior work [2], using default material and geometric parameters without inner constraints. Gravity is excluded, and symmetry is applied by fixing either one end cap or the central seam.

Two types of simulation sweeps were performed:

- Free-loading condition: Internal pressure $P_{\text{in}}$ is applied to the inner surface, increasing from 0 to 15 kPa in 0.5 kPa increments. This produces a displacement $x$ of the free end.

- Constrained condition: Using the final state from the free-loading step as the initial condition, an external force $F_{\text{ext}}$ is applied to the unfixed end cap. The force increases from 0 to 1.5 N in 0.05 N increments, counteracting the displacement while keeping $P_{\text{in}}$ fixed. The process is repeated for multiple fixed pressures.

The resulting dataset consists of paired actuation inputs, displacements, and reaction forces under both free and constrained loading. These data form the basis for constructing compact surrogate models.

## 1.2 Detailed Derivation of the Polynomial Surrogate

The actuator's joint force $\tau$ is defined from the derivative of the potential energy $U$ with respect to deformation $u$. With FEM results available only at discrete quasi-static states, this derivative can be approximated by finite differences:

$$\tau_i(p_i, u_i) \approx \frac{U_{i+1}(p_{i+1}, u_{i+1}) - U_i(p_i, u_i)}{u_{i+1} - u_i} \tag{S1}$$

where $i$ indexes the FEM step, and $(p_i, u_i)$ are the actuation–deformation pairs.

When the actuator is unconstrained, deformation varies with the applied actuation $p$. The equilibrium at each state requires that the actuation-induced force balances the elastic restoring force:

$$\tau_{a,f} + \tau_{e,f} = 0 \tag{S2}$$

with the elastic contribution evaluated as

$$\tau_{e,f} = \frac{\delta U_{e,f}}{\delta u} \tag{S3}$$

In the constrained case, the actuation level $p$ is fixed, while an external load $\tau_n$ is applied to affect deformation. The corresponding elastic force can be written as

$$\tau_{e,c}(p = P, u) = \tau_{e,f}(p = P, u_f) + \frac{\delta U_{e,c}(p = P, u)}{\delta u} \tag{S4}$$

where $u_f$ denotes the free-loading deformation under the same actuation. The equilibrium relation then becomes

$$\tau_{a,c} + \tau_{e,c} + \tau_n = 0 \tag{S5}$$

By further differentiating the force contributions with respect to $u$, the effective stiffness can be separated into an elastic part $k_e(u)$ and an actuation-dependent part $k_a(p)$. This leads to

$$\tau_n(p, u) = \tau_n(p, 0) - \int_0^u (k_e(u) + k_a(p)) du \tag{S6}$$

with $\tau_n(p, 0)$ determined from free-loading conditions.

Finally, each term $k_a(p)$, $k_e(u)$, and $\tau_n(p, 0)$ is approximated using low-order polynomial fits to FEM data. This produces a closed-form surrogate mapping $(p, u) \to \tau_n$, which is compact and suitable for fast evaluation in rigid-body-based simulators.

## 1.3 Example: Translational Bellow-SPA

To illustrate the surrogate construction and validate its viability, we applied the derivation to a translational bellow-SPA module. The FEM model of the module is shown in Figure S1a, and the strain energy and axial-displacement data are extracted under both free-loading and constrained conditions. Figure S1b further shows the calculated elastic, actuation, and external forces across simulation steps, in which the elastic force is derived using Eqs. S2 while the actuation force is obtained from the equilibrium conditions in Eqs. S3 and Eqs. S5. These results highlight the transition from free-loading to constrained conditions and motivate the separation of stiffness contributions, which we quantify as follows:

- Actuation stiffness $k_a(p)$: obtained from constrained FEM simulations at fixed pressures (5–15 kPa), then fitted with a first-order polynomial in $p$. The comparison in Figure S2a–b shows close agreement, with maximum relative error $\approx 7.7\%$.

- Elastic stiffness $k_e(u)$: obtained from constrained data as the slope of force–displacement curves, fitted linearly in $u$. Figure S2c shows a maximum relative error $\approx 4.1\%$.

- Offset term $F_n(p, 0)$: determined from free-loading FEM simulations (where $F_n(p, u_f) = 0$, fitted linearly in $p$ (Figure S2d, max relative error $\approx 0.1\%$).

Although simulations were performed across the full range of pressures (5–15 kPa), three representative cases (5, 10, and 15 kPa) are shown in Figure S2a&c for clarity, while regression and error analysis were performed over the complete dataset.

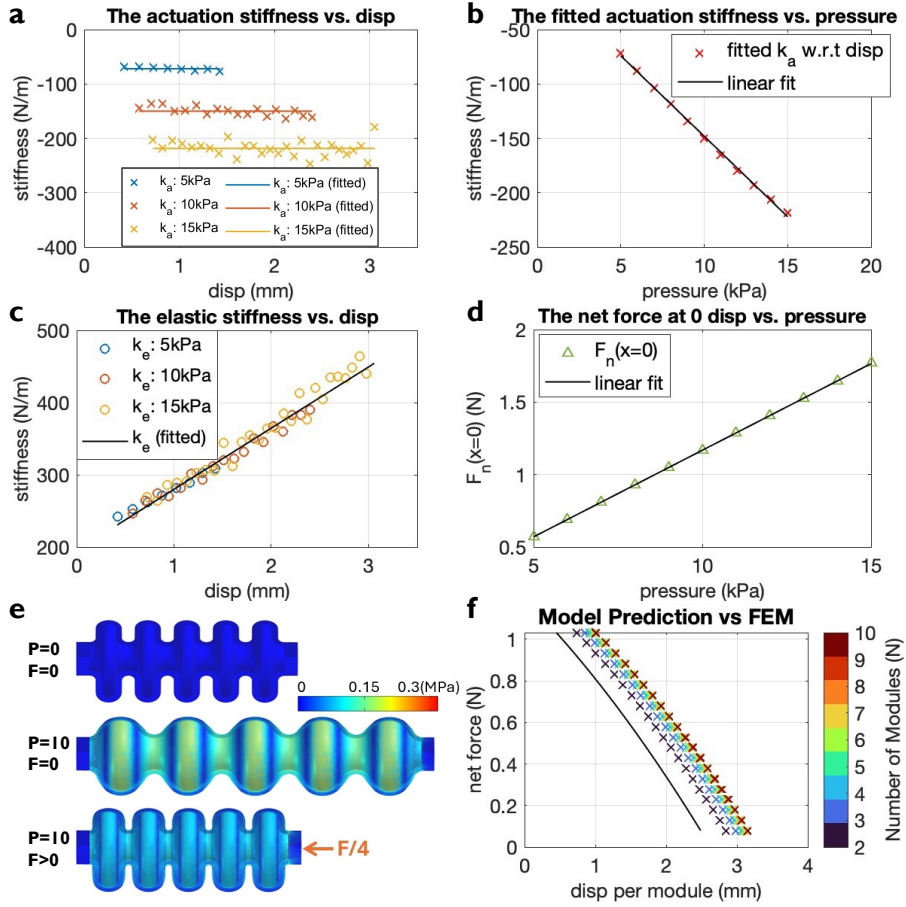

Figure S2: Polynomial-based surrogate modeling process and validation. (a–d) Polynomial fits of the actuation and elastic stiffness across different pressures and displacements. (e) FEM model of connected actuator modules. (f) Comparison between predictions from the polynomial surrogate model and the full FEM simulation for actuators composed of 2–10 modules.

These fits yield the closed-form expression

$$F_n(P, x) = -\int_0^x (m_a P + b_a + m_e x + b_e) dx + k_n P \tag{S7}$$

where coefficients are obtained from regression.

To further validate this formulation, we applied it to multi-module actuators. Figure S2e shows FEM simulations of actuators composed of 2–10 connected modules, each subjected to an internal pressure of 10 kPa. The surrogate model predicts the net force–displacement relationship by treating the full actuator as a series of identical modules, where the displacement scales with the number of modules while the force remains that of a single module.

Figure S2f compares FEM predictions with those from the surrogate model. Under the same applied pressure (10 kPa) and constrained end conditions, the surrogate consistently predicts smaller displacements than FEM for a given net force. The discrepancy increases with the number of modules, from about 16.1% error for 2 modules to 29.8% error for 10 modules, measured relative to the displacement range. Notably, the error growth slows as the number of modules increases, and the predictions converge toward a consistent offset.

This conservative underestimation arises from intermodule effects, such as accumulated compliance and boundary interactions, which are captured in FEM but not explicitly modeled in the surrogate's module-wise composition. Nevertheless, the surrogate accurately preserves the overall trend of the actuator's behavior, offering reliable tendency predictions with a beneficial safety margin for task-driven applications where ensuring feasibility is critical.

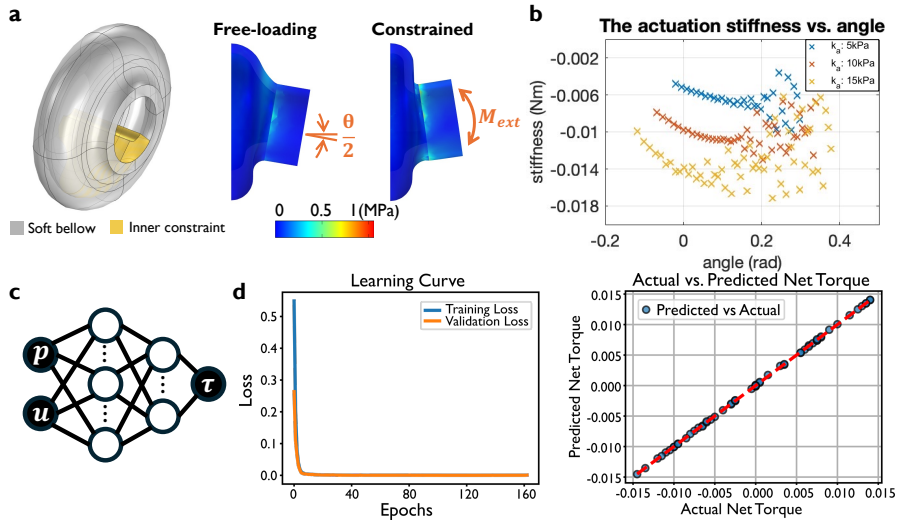

Figure S3: Surrogate modeling using small NNs. (a) FEM model of a bending bellow-SPA with an internal constraint. (b) Calculated actuation stiffness as a function of angular deflection under multiple pressure levels, obtained using the formulation in Section 3.2. (c) Architecture of the small feedforward NN. (d) Training and prediction results of the neural network surrogate.

## 1.4 Neural Network Surrogate Details

To illustrate the need for NN surrogates, we considered a bending bellow-SPA module with an internal constraint. Following the same modeling process as in Section 3.2, the deformation is defined as an angular deflection $\theta$, and the external force as a moment $M_{ext}$ (Figure S3a). The actuation stiffness of the default design is plotted against the modular angle for multiple pressures in Figure S3b.

The stiffness–angle curves exhibit strong nonlinearities, particularly at larger deflections, which low-order polynomial fits fail to capture. For instance, polynomial regressions of orders 1–3 yield maximum errors between 23.54% and 19.13% relative to the stiffness range, demonstrating the limitations of closed-form approximations in such cases.

To overcome this, we trained a small feed-forward NN to approximate the mapping $(p, \theta) \mapsto M_{ext}(p, \theta)$, as shown in Figure S3c. The network consists of two hidden layers (16 and 8 neurons, ReLU activations) and a linear output, trained using FEM data spanning free-loading and constrained conditions. Data are split into 80% training and 20% validation. The loss is defined as the mean squared error (MSE) between FEM-derived moments and network predictions, optimized with Adam (learning rate 0.005, batch size 16, 400 epochs with early stopping).

The trained network achieved an $R^2$ score above 0.999 on the validation set, substantially reducing errors compared with polynomial fits (Figure S3d). After training, the model remains sufficiently lightweight to be directly embedded in pseudo-rigid body model (PRBM)-based simulations in PyBullet, thereby enabling task-level simulation with nonlinear actuator behaviors.

## 2 Results and Discussion

This section complements Section 4 of the main text by providing supporting details on the simulation setup and optimization results.

### 2.1 Additional information of the Grasping Simulation

To determine appropriate PyBullet simulation parameters for the grasping task, an ablation study was performed on the simulation time step size and the joint damping coefficient, while keeping the actuation pressure, contact friction, and object properties fixed. For each tested parameter combination, a grasping simulation was executed and the dynamic responses of both the actuator and the object were evaluated.

Representative time histories of the actuator joint angles and joint torques are shown in Figure S4 to illustrate typical underdamped and overdamped behaviors. Very small damping values lead to pronounced oscillations in both

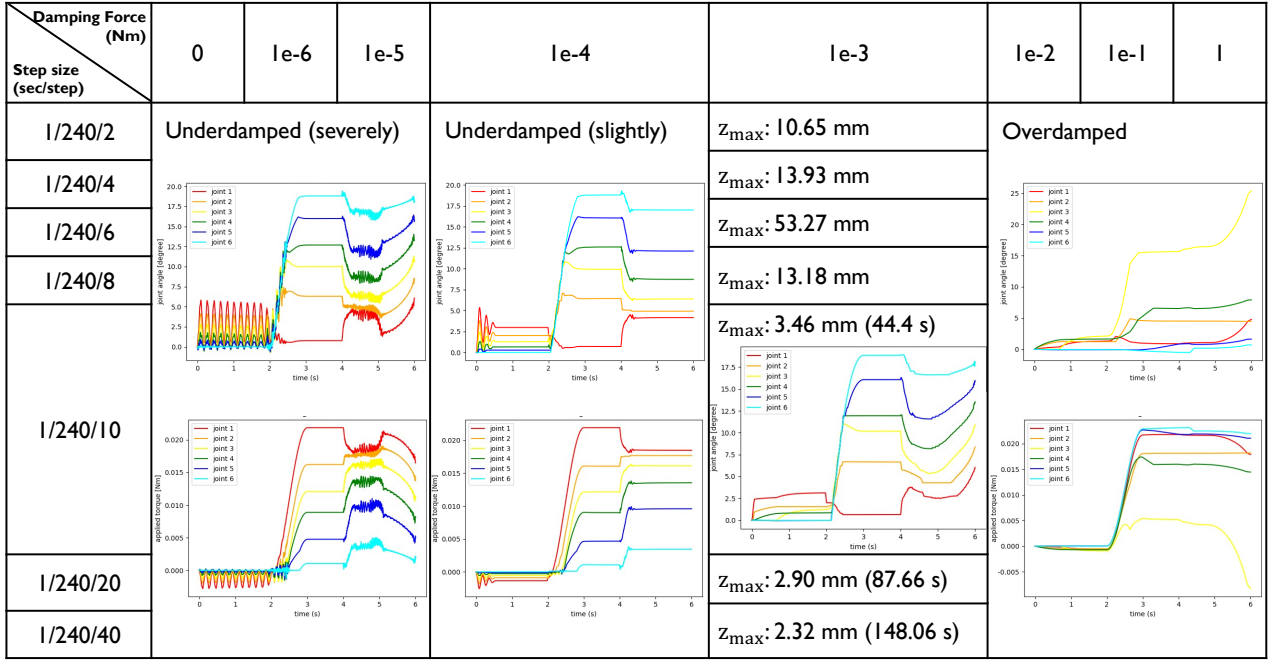

Figure S4: Ablation study of the PyBullet simulation settings for the grasping task. The effects of the simulation time step size and the joint damping coefficient are evaluated under fixed actuation pressure, contact friction, and object properties. Representative actuator joint angle and joint torque responses, together with the maximum vertical position of the object $z_{\max}$, are analyzed.

joint angles and joint torques, indicating underdamped responses, whereas excessively large damping suppresses joint motion and results in overdamped behavior with reduced effective actuation.

In addition, the maximum vertical position of the object, denoted as $z_{\max}$, is used to assess the effect of the simulation time step on the lifting performance. In Figure S4, the corresponding $z_{\max}$ values and wall-clock simulation times are explicitly reported for the cases with a joint damping coefficient of $10^{-3}$, thereby highlighting the trade-off between numerical accuracy and computational cost as the time step decreases. For this damping value, $z_{\max}$ increases and tends to converge when the time step reaches 1/2400 and below, while further reducing the time step mainly leads to increased simulation time.

Considering both numerical stability and computational efficiency, a simulation time step of 1/2400 and a joint damping coefficient of $10^{-3}$ were selected for all subsequent grasping simulations. This setting avoids both underdamped and overdamped behaviour while providing converged lifting performance with a reasonable simulation time.

To quantify the corresponding computational cost, we note that, using an Intel(R) Core(TM) i7-10875H CPU @ 2.30 GHz with 16.0 GB RAM, the execution time of a single simulation run ranged from approximately 10 to 60 s without visualization, depending on the complexity of the contact interactions.

## 2.2 Additional information of the Continuous Deformation Simulation

To determine suitable PyBullet simulation parameters for the helical actuator, we performed an ablation study on the simulation time step size and the joint damping coefficient. For each parameter combination, the actuator motion was simulated under identical loading conditions, and the actuator tip position (fifth ring) was analyzed after the transient response had settled. The simulated tip position was computed as the mean value over the last half cycle of the motion.

As shown in Figure S5a, the position error between the simulated and experimentally measured actuator tip varies significantly with both the time step size and the damping coefficient. Larger time steps generally lead to increased errors, while excessively small or large damping values also degrade accuracy. Figure S5b reports the standard deviation of the simulated tip position over the last half cycle, which reflects the numerical stability of the simulation. Insufficient damping results in large fluctuations, whereas overly strong damping suppresses motion and introduces bias.

Among all tested combinations, the parameter setting with a time step of 1/2400 and a joint damping coefficient of $10^{-4}$ achieves both low position error and low standard deviation. This setting therefore provides a favorable balance between simulation accuracy and numerical stability and is adopted for all subsequent helical actuator simulations in this work.

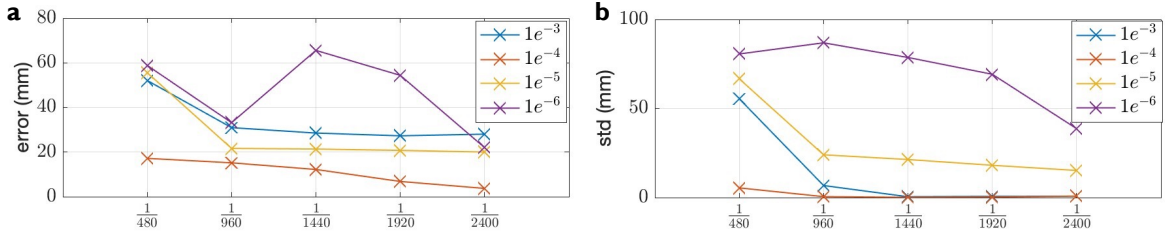

Figure S5: Ablation study of the PyBullet simulation settings for the helical actuator. The effects of the simulation time step size and the joint damping coefficient are evaluated at the fifth ring (actuator tip). (a) Position error between the experimentally measured actuator tip and the simulated tip position, where the simulated position is computed as the mean over the last half cycle after the motion has settled. (b) Standard deviation of the simulated tip position over the last half cycle.

To quantify the associated computational cost under this selected setting, we note that, under the same hardware configuration, the execution time of a single simulation run was approximately 10 s without visualization.

## 2.3 Details of the Task-Driven Soft Gripper Design with RL

From the learned distributions shown in Figure S6 and S7, the following representative values can be extracted for the optimized grasping strategy. The grasping height consistently converges to 7.5 mm, and the maximum pressure converges to 10 kPa across all tasks. The actuator inclination angle converges to approximately 0.314 rad. The optimized gripper morphology exhibits a strong convergence behavior: the average radius converges to 8 mm, the module length to 10 mm, the module number to 9, and the actuator number to 3.

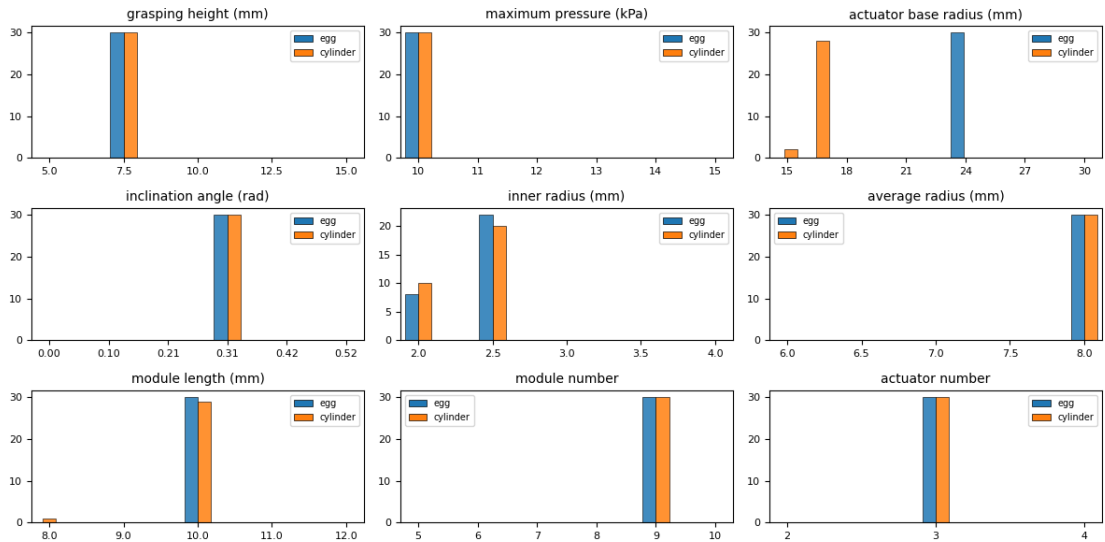

Figure S6: Distributions of the RL-generated design and control parameters for different object types. Histograms show the distributions of the optimized parameters for grasping an egg (black) and a cylinder (orange), including grasping height, maximum pressure, actuator base radius, inclination angle, inner radius, average radius, module length, module number, and actuator number. The y-axis represents the number of evaluation samples, while the x-axis corresponds to the parameter indicated in each subplot title. When two bars appear side by side, they share the same parameter value, defined by the center of the two bars; when only a single bar is present, its center indicates the corresponding parameter value.

6

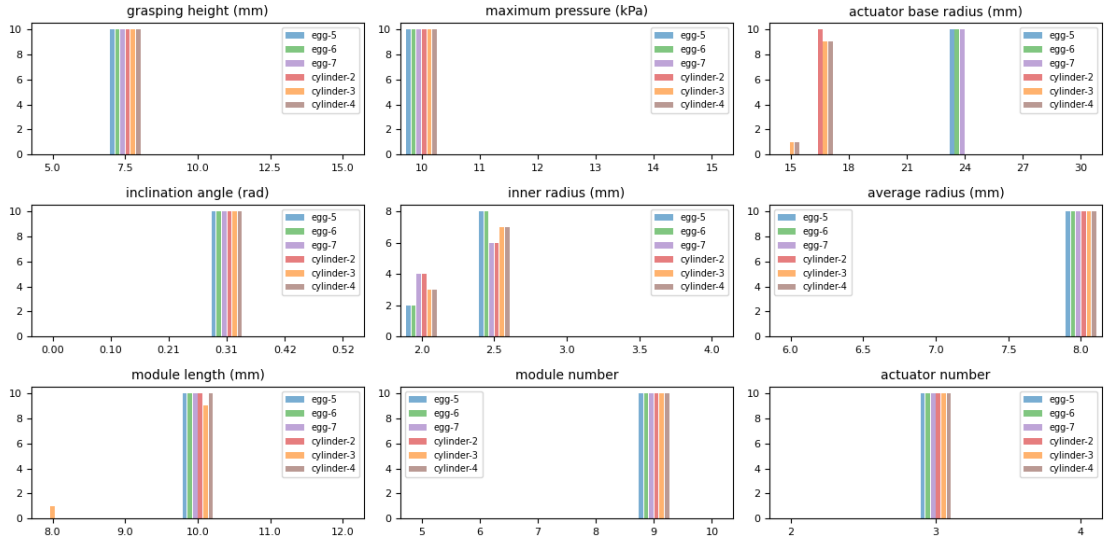

Figure S7: Distributions of the RL-generated design and control parameters across object types and weight conditions. Histograms show the optimized parameter distributions for grasping eggs and cylinders under different weight cases, including grasping height, maximum pressure, actuator base radius, inclination angle, inner radius, average radius, module length, module number, and actuator number. The y-axis represents the number of evaluation samples, while the x-axis corresponds to the parameter indicated in each subplot title. When bars appear side by side, they share the same parameter value, defined by the center of the bars; when only a single bar is present, its center indicates the corresponding parameter value.

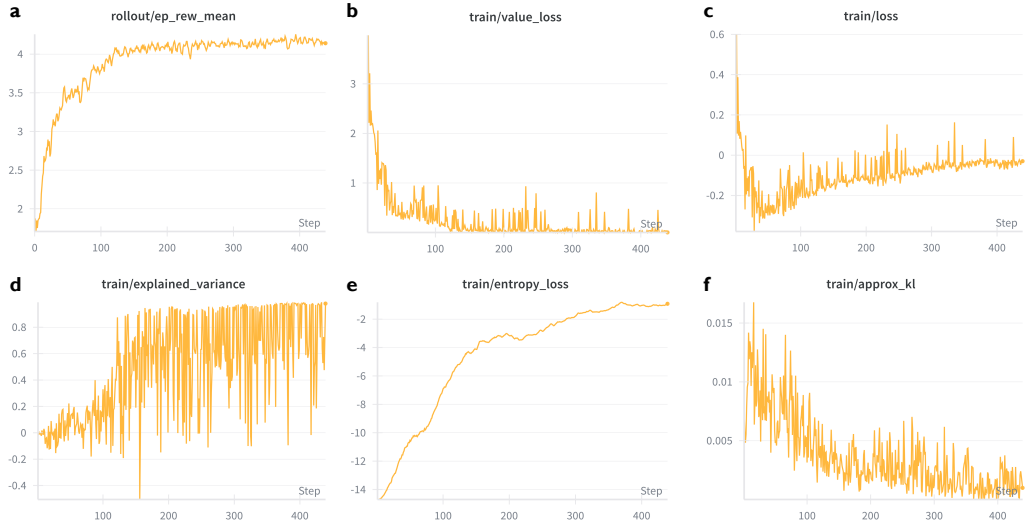

Figure S8: Training performance of the RL agent (PPO). (a) Mean episode reward, (b) value loss, (c) total training loss, (d) explained variance, (e) policy entropy, and (f) approximate Kullback–Leibler (KL) divergence, plotted over the training steps.

A task-dependent differentiation is observed only in the actuator base radius: the optimal value converges to approximately 24 mm for the egg and 17 mm for the cylinder, indicating a clear adaptation of the actuator placement to the object geometry. The inner radius exhibits a bimodal distribution around 2.0 mm and 2.5 mm. Since the distribution at 2.5 mm is more dominant and no clear task-dependent separation is observed, this value is selected as the final optimal design choice.

The task-driven design problem is solved using Proximal Policy Optimization (PPO) implemented with Stable-Baselines3. The agent is trained in a vectorized simulation environment consisting of four parallel environments

(DummyVecEnv), where each environment samples the object type (egg or cylinder) and weight randomly at reset. The policy network is a multilayer perceptron (MLP) operating under the compact observation mode.

The PPO hyperparameters are summarized as follows. The learning rate is set to $1 \times 10^{-4}$ to improve stability under task randomization. The discount factor is set to $\gamma = 0.9$ since each episode corresponds to a single grasp attempt. The number of rollout steps per update is n_steps = 8, with a batch size of 32. The clipping range is set to 0.3 to allow slightly larger policy updates. To encourage exploration during joint optimization of control and morphology, the entropy coefficient is set to 0.03. Generalized Advantage Estimation is used with $\lambda = 0.9$, and the value function loss coefficient is 0.3. Gradient updates are stabilized using a maximum gradient norm of 0.5, and a target KL-divergence of 0.05 is enforced. Advantage normalization is enabled to further smooth gradient updates.

The training curves are shown in Figure S8. As shown in Figure S8a, the mean episode reward increases rapidly during the early stage of training and gradually converges to a stable plateau, indicating successful policy learning. Figure S8b and Figure S8c show that both the value loss and total training loss decrease substantially and stabilize, demonstrating stable critic learning and policy updates. The explained variance of the value function (Figure S8d) increases and remains consistently high, indicating accurate value function approximation. The policy entropy (Figure S8e) gradually decreases as training progresses, reflecting the transition from exploration to exploitation. Meanwhile, the approximate KL-divergence (Figure S8f) remains well below the target threshold, confirming that PPO updates stay within the desired trust region.

These results confirm that the PPO agent achieves stable and well-converged training behavior under the proposed task-driven design formulation.

Table S1 reports a quantitative comparison between the baseline/default gripper used in Section 4.1.1 and the RL-optimized gripper obtained in Section 4.2.1. The default gripper is evaluated at actuation pressures of 15 kPa and 13 kPa, while the optimized gripper operates at the pressure selected by the RL policy (10 kPa). For each object and weight condition, grasping simulations are repeated in three groups of ten trials. Performance is evaluated using the grasp success rate (out of 10 trials) over three groups and the averaged simulated maximum object lifting position $z_{\max}$ computed across the same trials.

Table S1: Quantitative comparison between the baseline gripper and the RL-optimized gripper, evaluated using the grasp success rate and the averaged simulated maximum object lifting position $z_{\max}$.

| Object | Num of Extra Weights (Total Mass) | Default Success (15 kPa) | Default Success (13 kPa) | Optimized Success | Default $z_{\max}$ (15 kPa) [mm] | Default $z_{\max}$ (13 kPa) [mm] | Optimized $z_{\max}$ [mm] |
|---|---|---|---|---|---|---|---|
| Egg | 5 (150 g) | 10/10 × 3 | 10/10 × 3 | 10/10 × 3 | 56.10 | 53.18 | 66.39 |
| Egg | 6 (170 g) | 10/10 × 3 | 5/10, 7/10, 3/10 | 10/10 × 3 | 55.41 | 31.82 | 65.95 |
| Egg | 7 (190 g) | 10/10 × 3 | 1/10, 0/10, 0/10 | 10/10 × 3 | 54.65 | 6.91 | 65.32 |
| Cylinder | 2 (50 g) | 10/10 × 3 | 0/10 × 3 | 10/10 × 3 | 58.50 | 9.02 | 65.62 |
| Cylinder | 3 (70 g) | 0/10 × 3 | 0/10 × 3 | 10/10 × 3 | 2.14 | 0.23 | 63.43 |
| Cylinder | 4 (90 g) | 0/10 × 3 | 0/10 × 3 | 10/10 × 3 | 0.22 | 0.22 | 61.91 |

## 2.4 Details of the 3D Shape-Matching with CMA-ES

In this section, we report the optimal actuator design obtained from the Covariance Matrix Adaptation Evolution Strategy (CMA-ES)–based 3D shape-matching procedure described in Section 4.2.2. The optimized design uses a fixed inner radius $r = 3.0$ mm and wall thickness $t = 1.5$ mm, consistent with the surrogate training dataset. The structure is partitioned into $N = 4$ segments with module counts $n_i|_{i=1}^{4} = [6, 4, 6, 6]$. For each segment $i$, Table S2 lists the optimized average radius $R_i$, module length $l_i$, and inter-segment rotation angle $\phi_i$. The corresponding optimal actuation pressure is $P_{\max} = 11.81$ kPa.

Table S2: Optimized actuator design parameters for the 3D helix shape-matching task.

| Segment $i$ | $n_i$ | $R_i$ [mm] | $l_i$ [mm] | $\phi_i$ [deg] |
|---|---|---|---|---|
| 1 | 6 | 6.65 | 4.37 | 0.0 |
| 2 | 4 | 5.68 | 4.82 | 18.1 |
| 3 | 6 | 6.22 | 5.20 | 40.5 |
| 4 | 6 | 5.87 | 4.50 | 65.3 |


\end{document}